\documentclass{article}

\usepackage[dblblindworkshop, final]{neurips_2026}

\usepackage[utf8]{inputenc} 
\usepackage[T1]{fontenc}    
\usepackage{url}            
\usepackage{booktabs}       
\usepackage{amsfonts}       
\usepackage{nicefrac}       
\usepackage{microtype}      
\usepackage{xcolor}         
\usepackage{multirow}
\usepackage{microtype}
\usepackage{graphicx}
\usepackage{stfloats}
\usepackage{caption}
\usepackage{wrapfig}
\definecolor{mydarkblue}{rgb}{0,0.08,0.45}
\usepackage[ 
    colorlinks=true,
    linkcolor=mydarkblue,
    citecolor=mydarkblue,
    urlcolor=mydarkblue,
    filecolor=mydarkblue
]{hyperref}

\usepackage{float}
\usepackage{wrapfig}
\usepackage{graphicx}  

\usepackage{amsmath}
\usepackage{amssymb}   

\usepackage{tikz}
\usetikzlibrary{arrows.meta, shapes.geometric, positioning, calc}
\usetikzlibrary{arrows.meta, positioning, fit, backgrounds, calc, shapes.geometric}

\usepackage{pifont}
\newcommand{\xmark}{\ding{55}}

\definecolor{myblue}{RGB}{0, 0, 255}
\definecolor{myred}{RGB}{200, 0, 0}
\definecolor{mygreen}{RGB}{0, 125, 0}

\usepackage{mathtools}
\usepackage{amsthm}
\usepackage[capitalize,noabbrev]{cleveref}
\usepackage{color,soul} 
\usepackage{siunitx}
\usepackage{siunitx}
\usepackage{subcaption}

\title{Manifold-Constrained Hyper-Connections for Parameter-Efficient Finetuning}
\workshoptitle{AXIOM: Foundations of Efficient Deep Learning}

\author{%
  Valentijn ~Oldenburg\thanks{\raggedright Correspondence to:\;
    \texttt{valentijn.oldenburg@student.uva.nl},\;
    \texttt{floris.de.kam@student.uva.nl},\;
    \texttt{bente.zuijdam@student.uva.nl}.}\\
  University of Amsterdam\\
   \And
   Floris ~de Kam\footnotemark[1] \\
   University of Amsterdam\\
   \And
   Bente ~Zuijdam\footnotemark[1] \\
   University of Amsterdam \\
   \AND
   Lieve ~Eberson \\
   University of Amsterdam \\
   \And
   Nicky ~van Zutphen\\
   University of Amsterdam  \\
     \And
  Stef ~de Wildt \\
  University of Amsterdam  \\
     \AND
   Ivo ~Verhoeven \\
  University of Amsterdam \\
     \And
  Cees ~Snoek \\
  University of Amsterdam \\
}

\begin{document}
\maketitle
\begin{abstract}
Finetuning methods for foundation models usually change weights, prompts, or hidden states, while leaving the residual topology fixed. We ask whether residual topology itself can become a finetuning object. To study this, we adapt manifold-constrained hyper-connections (mHC), recently introduced for pre-training, to frozen-backbone finetuning. mHC turns a Transformer into an input-dependent multi-stream residual architecture, routing representations through multiple streams at every sub-layer. Across mHC variants, we find that dynamic residual routing can finetune Transformers, but that its role differs from pre-training: by preserving stream mixing and learning only how sub-layers access streams, loss is improved and trainable parameters are reduced. Overall, our results identify residual routing as a promising architectural axis for efficient finetuning of foundation models.
\end{abstract}
\section{Introduction}\label{sec:intro}
Residual connections define the topology through which information flows over layers in deep neural networks \citep{veit2016residual, bicici2024residual, xu2025development}.
Nevertheless, most finetuning methods adapt weights, prompts, or activations, and leave this residual topology fixed \citep{han2024parameter}.
Originally developed for pre-training, hyper-connections (HC; \citet{zhu2025hyperconnections}) generalise the single residual stream to multiple streams that learn to interact.
Manifold-constrained hyper-connections (mHC; \citet{xie2026mhcmanifoldconstrainedhyperconnections}) constrain the residual mixing matrix onto a norm-preserving manifold, ensuring stable mixing in deep models, and were shown to produce complex residual routing with limited additional overhead in pre-training.
However, its potential for finetuning remains unexplored.

In this paper, we study mHC as an architectural finetuning mechanism for pre-trained Transformers.
We wrap each Transformer sub-layer with an mHC module that turns the fixed residual topology into an adaptive object, learning how representations are read from and written to multiple residual streams.
Our contributions are two-fold: (\emph{i}) we introduce residual routing as a novel parameter-efficient finetuning (PEFT) axis through mHC and (\emph{ii}) show that mHC behaves differently in finetuning than in pre-training: learned residual mixing moves toward identity and fixing it to identity improves loss and reduces trainable parameters.
Thus, in finetuning, mHC works best by fixing stream mixing and learning only how sub-layers access the residual streams.
Our code is available on \href{https://github.com/valentijn7/mHC_PEFT.git}{GitHub}.
\section{Background: HC and mHC}\label{sec:background}

HC generalise a single residual stream by maintaining $n$ streams,
$\mathbf{Z}^{l-1}=[\mathbf{z}^{l-1}_{1},\ldots,\mathbf{z}^{l-1}_{n}]^\top\in\mathbb{R}^{n\times d}$.
Before sub-layer $l$ with transformation $f^l$, HC reads $\mathbf{x}^{l}=(\mathbf{h}^{l}_{\mathrm{pre}})^\top\mathbf{Z}^{l-1}$ and writes the output back as
\begin{equation}\label{eq:hc_update_new}
    \mathbf{Z}^{l}
    =
    \mathbf{H}^{l}_{\mathrm{res}}\mathbf{Z}^{l-1}
    +
    \mathbf{h}^{l}_{\mathrm{post}} \otimes f^{l}(\mathbf{x}^{l}).
\end{equation}
Here, $\mathbf{h}^{l}_{\mathrm{pre}}\in\mathbb{R}^{n}$ reads from the streams, $\mathbf{H}^{l}_{\mathrm{res}}\in\mathbb{R}^{n\times n}$ mixes them, and $\mathbf{h}^{l}_{\mathrm{post}}\in\mathbb{R}^{n}$ writes the sub-layer output.
An unconstrained $\mathbf{H}^{l}_{\mathrm{res}}$, however, can destabilise residual streams over depth.

mHC \citep{xie2026mhcmanifoldconstrainedhyperconnections} constrains $\mathbf{H}^{l}_{\mathrm{res}}$ to be doubly stochastic through Sinkhorn normalisation \citep{sinkhorn1967concerning}, i.e.\ to lie in the Birkhoff polytope $\mathcal{B}_{n}$ \citep{birkhoff1946tres}.
This lets streams exchange information while preventing routing from freely growing or shrinking the residual signal.
With \emph{static} mHC, each sub-layer directly learns only the routing logits themselves: $n^2$ for $\mathbf{H}^{l}_{\mathrm{res}}$, and $n$ for $\mathbf{h}^{l}_{\mathrm{pre}}$ and $\mathbf{h}^{l}_{\mathrm{post}}$, giving $n^2+2n$ logits.
In contrast, \emph{dynamic} mHC predicts these same $n^2+2n$ routing logits from the current residual streams, requiring $nd(n^2+2n)$ projection weights, $n^2+2n$ biases, and $3$ gates per sub-layer.
It is thus more expressive, but no longer has the extremely small parameter count of the static variant.
These counts measure trainable parameters only: having $n$ residual streams also adds activation-memory and throughput costs, which we report in Appendix~\ref{ap:efficiency}.

Two recent variants, mHC-lite \citep{yang2026mhclitedontneed20} and KromHC \citep{zhou2026kromhcmanifoldconstrainedhyperconnectionskroneckerproduct}, parameterise $\mathbf{H}^{l}_{\mathrm{res}}$ more cheaply through mixtures of permutation matrices and Kronecker products of small doubly stochastic factors, resp., while keeping the read/write structure of Eq.~\eqref{eq:hc_update_new}; details in Appendix~\ref{ap:background_full}.
\section{Methods}\label{sec:method}

We build an mHC module that wraps the sub-layers of a pre-trained Transformer, which lets us ask which parts of the residual topology finetuning re-uses.
As the public code targets pre-training, we implement all six variants ourselves in PyTorch \citep{PyTorch}, i.e.\ a static and a dynamic version of original mHC, mHC-lite, and KromHC.
We use OLMo-2-0425-1B as main backbone and OLMo-2-1124-7B \citep{olmo20252olmo2furious} for scaling experiments.
The module only assumes residual blocks, so it is not specific to text data.
Appendix~\ref{ap:PEFT_baselines} describes the used PEFT baselines.

\begin{figure*}
\vspace{1em}
\centering
\resizebox{\textwidth}{!}{%
\begin{tikzpicture}[
    every node/.style={font=\Large},
    box/.style={
        rectangle, rounded corners=4pt,
        draw=black!40, line width=0.45pt,
        align=center, minimum height=0.9cm
    },
    iobox/.style={
        box, fill=gray!12, draw=gray!40,
        minimum width=3.0cm
    },
    streambox/.style={
        box, fill=teal!15, draw=teal!50,
        minimum width=1.8cm, minimum height=0.72cm
    },
    gatingbox/.style={
        box, fill=violet!12, draw=violet!50,
        minimum width=4.3cm
    },
    frozenbox/.style={
        box, fill=gray!8, draw=gray!55, dashed,
        minimum width=4.8cm, minimum height=1.05cm
    },
    routingbox/.style={
        box, fill=orange!15, draw=orange!50,
        minimum width=5.0cm, minimum height=3.0cm
    },
    readoutbox/.style={
        box, fill=blue!8, draw=blue!35,
        minimum width=3.8cm, minimum height=1.25cm
    },
    sumcircle/.style={
        circle, draw=black!45, fill=white,
        minimum size=5.5mm, inner sep=0pt
    },
    arr/.style={
        -{Stealth[length=4pt,width=3pt]},
        line width=0.55pt, draw=black!65
    },
    lab/.style={
        font=\large, text=black!60, align=center
    }
]

\node[iobox] (input) at (0,0)
    {$\mathbf{x} \in \mathbb{R}^{B \times T \times d}$};

\node[streambox] (s1) at (3.4, 1.5) {$X^{(1)}$};
\node[streambox] (s2) at (3.4, 0.5) {$X^{(2)}$};
\node[streambox] (s3) at (3.4,-0.5) {$X^{(3)}$};
\node[streambox] (s4) at (3.4,-1.5) {$X^{(4)}$};

\node[lab] at (3.70,2.35) {$n=4$ streams};

\foreach \s in {s1,s2,s3,s4}{
    \draw[arr] (input.east) -- ++(0.7,0) |- (\s.west);
}

\node[gatingbox] (hpre) at (9.0, 3.6)
    {$\mathbf{h}_{\mathrm{pre}}$};

\node[frozenbox] (branch) at (13.8, 3.6)
    {Attention / MLP};

\node[gatingbox] (hpost) at (18.6, 3.6)
    {$\mathbf{h}_{\mathrm{post}}$};

\coordinate (hpin1) at ($(hpre.south west)!0.16!(hpre.south east)$);
\coordinate (hpin2) at ($(hpre.south west)!0.38!(hpre.south east)$);
\coordinate (hpin3) at ($(hpre.south west)!0.62!(hpre.south east)$);
\coordinate (hpin4) at ($(hpre.south west)!0.84!(hpre.south east)$);

\draw[arr] (s1.east) -- (hpin1);
\draw[arr] (s2.east) -- (hpin2);
\draw[arr] (s3.east) -- (hpin3);
\draw[arr] (s4.east) -- (hpin4);

\draw[arr] (hpre.east) -- (branch.west);
\draw[arr] (branch.east) -- (hpost.west);

\node[routingbox] (hres) at (13.8, 0.0)
    {$\mathbf{H}_{\mathrm{res}}$ residual routing\\[2pt]
     cross-stream mixing \\ \large Sinkhorn$(n \times n)$};

\coordinate (hin1) at ($(hres.west)+(0, 1.10)$);
\coordinate (hin2) at ($(hres.west)+(0, 0.37)$);
\coordinate (hin3) at ($(hres.west)+(0,-0.37)$);
\coordinate (hin4) at ($(hres.west)+(0,-1.10)$);

\draw[arr] (s1.east) -- (hin1);
\draw[arr] (s2.east) -- (hin2);
\draw[arr] (s3.east) -- (hin3);
\draw[arr] (s4.east) -- (hin4);

\coordinate (hout1) at ($(hres.east)+(0, 1.10)$);
\coordinate (hout2) at ($(hres.east)+(0, 0.37)$);
\coordinate (hout3) at ($(hres.east)+(0,-0.37)$);
\coordinate (hout4) at ($(hres.east)+(0,-1.10)$);

\node[sumcircle] (m1) at (22.1, 1.5) {$+$};
\node[sumcircle] (m2) at (22.1, 0.5) {$+$};
\node[sumcircle] (m3) at (22.1,-0.5) {$+$};
\node[sumcircle] (m4) at (22.1,-1.5) {$+$};

\draw[arr] (hout1) -- (m1.west);
\draw[arr] (hout2) -- (m2.west);
\draw[arr] (hout3) -- (m3.west);
\draw[arr] (hout4) -- (m4.west);

\coordinate (hpout4) at ($(hpost.south west)!0.16!(hpost.south east)$);
\coordinate (hpout3) at ($(hpost.south west)!0.38!(hpost.south east)$);
\coordinate (hpout2) at ($(hpost.south west)!0.62!(hpost.south east)$);
\coordinate (hpout1) at ($(hpost.south west)!0.84!(hpost.south east)$);

\draw[arr] (hpout1) -- (m1.north);
\draw[arr] (hpout2) -- (m2.north);
\draw[arr] (hpout3) -- (m3.north);
\draw[arr] (hpout4) -- (m4.north);

\node[streambox] (u1) at (24.7, 1.5) {$X_{\mathrm{new}}^{(1)}$};
\node[streambox] (u2) at (24.7, 0.5) {$X_{\mathrm{new}}^{(2)}$};
\node[streambox] (u3) at (24.7,-0.5) {$X_{\mathrm{new}}^{(3)}$};
\node[streambox] (u4) at (24.7,-1.5) {$X_{\mathrm{new}}^{(4)}$};

\draw[arr] (m1.east) -- (u1.west);
\draw[arr] (m2.east) -- (u2.west);
\draw[arr] (m3.east) -- (u3.west);
\draw[arr] (m4.east) -- (u4.west);

\node[readoutbox] (readout) at (29.5, 0)
    {Softmax readout\\[2pt]
     \large $(B,T,n,d)\to(B,T,d)$};

\coordinate (rin1) at ($(readout.west)+(0, 0.60)$);
\coordinate (rin2) at ($(readout.west)+(0, 0.25)$);
\coordinate (rin3) at ($(readout.west)+(0,-0.25)$);
\coordinate (rin4) at ($(readout.west)+(0,-0.60)$);

\draw[arr] (u1.east) -- (rin1);
\draw[arr] (u2.east) -- (rin2);
\draw[arr] (u3.east) -- (rin3);
\draw[arr] (u4.east) -- (rin4);

\draw[arr] (readout.east) -- ++(1.0,0);

\node[lab] at ($(hpre.north)+(0,0.20)$)  {$n$ params};
\node[lab] at ($(hpost.north)+(0,0.20)$) {$n$ params};
\node[lab] at ($(hres.north)+(0,0.42)$)  {$n^2$ params};

\node[
    draw=black!55,
    dashed,
    rounded corners=6pt,
    inner sep=20pt,
    fit=(s1)(s4)(hpre)(branch)(hpost)(hres)(u1)(u4),
    label={[font=\large\bfseries, text=black!65]above:
      x32}
] (wrapper) {};

\node at (28.5, 5.0) (leganchor) {};

\node[streambox, minimum width=0.4cm, minimum height=0.6cm,
      below=0cm of leganchor] (lt) {};
\node[right=0.12cm of lt, lab] (ltxt) {streams};

\node[gatingbox, minimum width=0.4cm, minimum height=0.6cm,
      below=1.0cm of leganchor] (lg) {};
\node[right=0.12cm of lg, lab] (gtxt) {gating (trained)};

\node[routingbox, minimum width=0.4cm, minimum height=0.6cm,
      below=2.0cm of leganchor] (lr) {};
\node[right=0.12cm of lr, lab] {routing (trained)};

\end{tikzpicture}%
}
\caption{
mHC-wrapped Transformer block.\ The $n$ streams learn how each frozen sub-layer reads from, mixes, and writes back to them.\ $B$ is batch size, $T$ sequence length, and $d$ embedding dim.
}
\label{fig:wrapvisual}
\vspace{0.5em}
\end{figure*}

\textbf{Implementing mHC wrappers.}\quad
We wrap each attention and MLP sub-layer with an mHC module that changes only the residual routing around the frozen branch and leaves the block's input-output interface intact, see Figure~\ref{fig:wrapvisual}.
Any change in loss can then be attributed to routing.
Internally, the hidden representation is lifted to $n$ residual streams and reduced back to one representation after the final layer.
Both are initialised close to the pre-trained behaviour, so finetuning starts from the pre-trained optimum.
Appendix~\ref{ap:method_full} details initialisation, symmetry breaking, and parameter counts.

\textbf{Training and evaluation.}\quad
We finetune on the Tulu-3 SFT OLMo mixture \citep{lambert2024tulu} used in prior finetuning studies, e.g., \citet{sun2025amurocharanalyzingrelationship,springer2025overtrainedlanguagemodelsharder}.
Appendix~\ref{sec:method_data} describes further preprocessing.
We evaluate held-out next-token loss, eight downstream benchmarks, trainable parameter count, and efficiency costs, since loss says little about task ability and parameter count says little about the cost of a structural adapter.
Following \citet{xie2026mhcmanifoldconstrainedhyperconnections}, we benchmark with the \texttt{lm-eval} harness \citep{eval-harness,biderman2024lessonstrenchesreproducibleevaluation}, with specifics in Appendix~\ref{ap:benchmark_details}.
\section{Experiments}\label{sec:experiments}

\textbf{Identity preservation.}\quad
Our first experiment asks which part of mHC is used during finetuning: the learned residual mixing $\mathbf{H}^{l}_{\mathrm{res}}$, or the read/write vectors $\mathbf{h}^{l}_{\mathrm{pre}}$ and $\mathbf{h}^{l}_{\mathrm{post}}$.
To fix one parameterisation, we first compare the six mHC variants (static/dynamic for Sinkhorn mHC/mHC-lite/KromHC) with $n=4$ for \num{3000} steps in Table~\ref{tab:mhc_variant_comparison}.
Static KromHC trains only \num{384} parameters but does reduce test loss from \num{2.00} to \num{1.77}.
Dynamic KromHC performs best while being the most \mbox{(parameter-)efficient} dynamic variant, so we use it for further experiments.
Next, we vary $n\in\{2,4,8,16\}$ and compare the standard learned residual mixer $\mathbf{H}^{l}_{\mathrm{res}}$ with an identity-preserving one, i.e.\ $\mathbf{H}^{l}_{\mathrm{res}}=\mathbf{I}_n$.

Table~\ref{tab:kromhc_expansion_identity} shows two patterns.
First, increasing $n$ improves loss, but with diminishing returns and fast-growing parameter count.
Second, fixing $\mathbf{H}^{l}_{\mathrm{res}}=\mathbf{I}_n$ improves test loss for every $n$ while removing residual-mixing parameters, and more so for larger $n$.
This suggests that, in finetuning, the useful capacity comes mainly from the read/write vectors $\mathbf{h}^{l}_{\mathrm{pre}}$ and $\mathbf{h}^{l}_{\mathrm{post}}$, while fixing $\mathbf{H}^{l}_{\mathrm{res}}=\mathbf{I}_n$ is beneficial.
Figure~\ref{fig:residual_mixing_l13} supports this, showing some layers spontaneously revert to near-identity mixing during training.
This differs from pre-training, where learned residual mixing is an important part of the mechanism \citep{xie2026mhcmanifoldconstrainedhyperconnections}.
The practical conclusion is therefore architectural: in finetuning, mHC should preserve the residual topology and adapt only how modules read from and write to it.
Interpretability experiments, ablations, and alternative initialisations in Appendix~\ref{ap:experiments} all support this interpretation.
From here on, we refer to dynamic KromHC with $\mathbf{H}^{l}_{\mathrm{res}}=\mathbf{I}_n$ as mHC$_{\mathrm{identity}}$.

\begin{table*}[t]
\centering
\vspace{1em}

\begin{minipage}[t]{0.48\textwidth}
\vspace{0pt}
\centering
\caption{Comparison of mHC variants. KromHC performs best and is most (parameter-)efficient.}
\label{tab:mhc_variant_comparison}
\vspace{-0.6em}
\footnotesize
\setlength{\tabcolsep}{2.5pt}
\renewcommand{\arraystretch}{0.9}
\begin{tabular}{lccc}
\toprule
Method & \# params& $\mathcal{L}_\mathrm{train}$ & $\mathcal{L}_\mathrm{test}$ \\
\midrule
OLMo-2-1B & -- & -- & 2.00 \\
\midrule
static mHC & 772 & 1.85 & 1.79 \\
static mHC-lite & 1024 & 1.83 & 1.77 \\
static KromHC & 384 & 1.83 & 1.77 \\
\midrule
dynamic mHC & $\qty{6.55}{\text{M}}$ & 1.48 & 1.39 \\
dynamic mHC-lite & $\qty{8.65}{\text{M}}$ & 1.49 & 1.40 \\
dynamic KromHC & $\qty{3.41}{\text{M}}$ & 1.42 & 1.35 \\
\bottomrule
\end{tabular}
\end{minipage}
\hfill
\begin{minipage}[t]{0.48\textwidth}
\vspace{0pt}
\centering
\includegraphics[width=\textwidth]{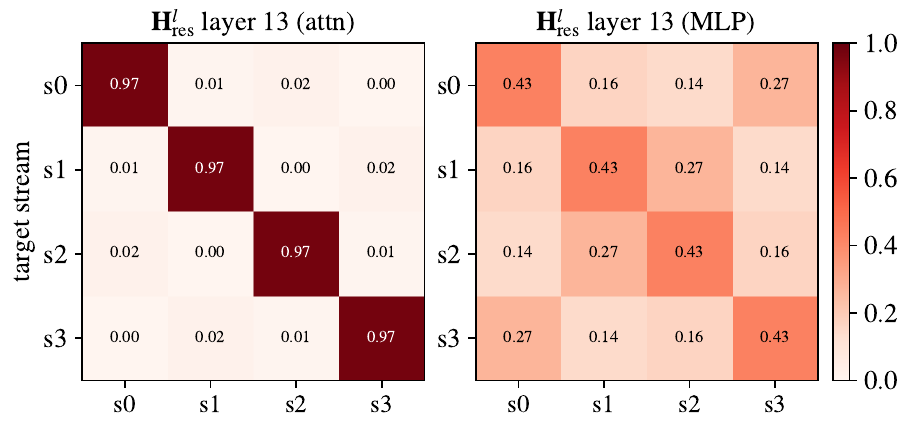}
\vspace{-1.5em}
\captionof{figure}{$\mathbf{H}^{l}_{\mathrm{res}}$ within static mHC for MLP and attention layer 13, which converges to $\mathbf{H}^{l}_{\mathrm{res}} \approx \mathbf{I}$.
}
\label{fig:residual_mixing_l13}
\vspace{-0.5em}
\end{minipage}

\vspace{-1em}
\end{table*}

\begin{table}[t]
\vspace{-0em}
\centering
\caption{
Dynamic KromHC on OLMo-2-1B with learned residual mixing and with fixed $\mathbf{H}^{l}_{\mathrm{res}}=\mathbf{I}_n$.
$\Delta\mathcal{L}_{\mathrm{test}}$ is fixed minus learned, so lower is better.
Fixing $\mathbf{H}^{l}_{\mathrm{res}}=\mathbf{I}_n$ improves test loss for every $n$.}
\vspace{0.1em}
\label{tab:kromhc_expansion_identity}
\setlength{\tabcolsep}{4pt}
\footnotesize
\begin{tabular}{lccc@{\hskip 1.2em}ccc@{\hskip 1.2em}cc}
\toprule
& \multicolumn{3}{c}{Learned $\mathbf{H}^{l}_{\mathrm{res}}$}
& \multicolumn{3}{c}{Fixed $\mathbf{H}^{l}_{\mathrm{res}}=\mathbf{I}_n$}
& \multicolumn{2}{c}{Effect of fixed $\mathbf{H}^{l}_{\mathrm{res}}=\mathbf{I}_n$} \\
\cmidrule(lr){2-4} \cmidrule(lr){5-7} \cmidrule(lr){8-9}
$n$ & \# params & $\mathcal{L}_\mathrm{train}$ & $\mathcal{L}_\mathrm{test}$
    & \# params & $\mathcal{L}_\mathrm{train}$ & $\mathcal{L}_\mathrm{test}$
    & \# params \underline{saved} & $\Delta\mathcal{L}_\mathrm{test}$ \\
\midrule
(base OLMo-2-1B)       & \num{0} & -- & 2.00 & \num{0} & -- & 2.00 & -- & -- \\
\midrule
\num{2}  & \qty{918}{\text{K}}  & 1.45 & 1.37 & \qty{655}{\text{K}}  & 1.42 & 1.36 & \qty{262}{\text{K}}  & $-0.01$ \\
\num{4}  & \qty{3.41}{\text{M}} & 1.42 & 1.35 & \qty{2.36}{\text{M}} & 1.39 & 1.33 & \qty{1.05}{\text{M}} & $-0.02$ \\
\num{8}  & \qty{12.1}{\text{M}} & 1.40 & 1.33 & \qty{8.91}{\text{M}} & 1.37 & 1.31 & \qty{3.15}{\text{M}} & $-0.02$ \\
\num{16} & \qty{43.0}{\text{M}} & 1.39 & 1.32 & \qty{34.6}{\text{M}} & 1.33 & 1.27 & \qty{8.39}{\text{M}} & $-0.05$ \\
\bottomrule
\end{tabular}
\end{table}

\begin{wraptable}{r}{0.43\textwidth}
\vspace{-1.2em}
\centering
\caption{Total training cost/time for \num{20000} steps on OLMo-2-7B. Energy is GPU-only.
}
\vspace{-0.5em}
\label{tab:cost_short}
\scriptsize
\setlength{\tabcolsep}{2.1pt}
\renewcommand{\arraystretch}{0.9}
\begin{tabular}{llccc}
\toprule
GPU & Method & \# params & Time & GPU energy \\
\midrule
\multirow{2}{*}{A100} 
& LoRA$_{r=1}$  & \qty{2.50}{\text{M}} & \qty{57.8}{\hour} & \qty{20.1}{\kWh} \\
& LoRA$_{r=14}$ & \qty{35.0}{\text{M}} & \qty{59.7}{\hour} & \qty{19.9}{\kWh} \\
\midrule
\multirow{2}{*}{H100} 
& mHC$_{\mathrm{identity},n=2}$ & \qty{2.62}{\text{M}} & \qty{37.8}{\hour} & \qty{19.5}{\kWh} \\
& mHC$_{\mathrm{identity},n=4}$ & \qty{9.44}{\text{M}} & \qty{48.8}{\hour} & \qty{23.7}{\kWh} \\
\bottomrule
\end{tabular}
\vspace{-1em}
\end{wraptable}

\textbf{Efficiency beyond parameters.}\quad
To test whether parameter count reflects the practical cost of a structural adapter, we measure wall-clock time and GPU energy for \num{20000}-step OLMo-2-7B runs, scaling LoRA by rank $r$ and mHC by streams $n$, see Table~\ref{tab:cost_short}.
Since both ran on different GPU types, we only compare within each type.
For LoRA, going from $r=1$ to $r=14$ barely changes training time; for mHC, going from $n=2$ to $n=4$ increases time and energy more clearly.
This shows why parameter count is insufficient as efficiency indicator: while the read-in gate keeps backbone computation and the key-value cache independent of $n$, stream expansion changes the residual state from $(B,T,d)$ to $(B,T,n,d)$, which must be routed and stored during training.
Thus, mHC is parameter-efficient, but its practical cost depends on stream expansion, not just on trainable parameter count.
More broadly, residual routing shifts the efficiency question from parameter- to architectural efficiency: mHC lifts a standard Transformer into a multi-stream architecture whose cost depends on the expanded residual topology.

\begin{table*}[t]
\centering
\caption{
Downstream benchmark results on OLMo-2-1B.
Best scores are \textcolor{myblue}{\underline{underlined}} separately for single methods and combinations.
N.Acc.\ is normalised accuracy; EM is exact match. No method dominates, performance is task-dependent and not monotonic in parameter count.
}
\vspace{-0.4em}
\label{tab:benchmark_results_1b}
\setlength{\tabcolsep}{1.15pt}
\renewcommand{\arraystretch}{0.88}
\resizebox{\textwidth}{!}{
\begin{tabular}{lllcccccccc}
\toprule
Type & Method & \# trainable params & BBH & DROP & GSM8K & HellaSwag & MATH & MMLU & PIQA & TriviaQA \\
(Metric) & & & (EM) & (F1) & (EM) & (N.Acc.) & (EM) & (Acc.) & (Acc.) & (EM) \\
\# shots & & & 3-shot & 3-shot & 8-shot & 10-shot & 4-shot & 5-shot & 0-shot & 5-shot \\
\midrule
Base & OLMo-2-1B & 0 
& 20.73 & 2.22 & 27.60 & 49.58 
& \textcolor{myblue}{\underline{6.14}} 
& 31.41 & 75.52 & 27.20 \\
\midrule
\multirow{10}{*}{Single} 
& prompt tuning & \qty{41.0}{\text{K}} 
& 29.04 & 3.20 & 29.57 & 49.87 & 5.58 
& \textcolor{myblue}{\underline{31.89}} 
& 75.95 & 25.85 \\
& $(\mathrm{IA})^3$ & \qty{328}{\text{K}} 
& \textcolor{myblue}{\underline{29.30}} 
& 2.56 & 28.73 & 49.38 & 5.74 & 26.29 & 75.19 & 25.59 \\
& VeRA & \qty{427}{\text{K}} 
& 27.85 & 2.66 & 29.57 & 49.31 & 5.14 & 29.63 & 74.54 & 25.94 \\
& LoRA$_{r=8}$ & \qty{6.03}{\text{M}} 
& 26.66 
& \textcolor{myblue}{\underline{5.41}} 
& \textcolor{myblue}{\underline{31.39}} 
& 50.06 & 1.50 & 23.73 & 74.70 & 0.06 \\
& LoRA$_{r=9}$ & \qty{6.78}{\text{M}} 
& 26.66 & 5.30 & 27.98 & 49.30 & 2.62 & 24.95 & 75.03 & 0.03 \\
& layer tuning & \qty{67.1}{\text{M}} 
& 25.19 & 0.43 & 14.86 
& \textcolor{myblue}{\underline{50.26}} 
& 0.44 & 28.94 & 74.92 & 0.04 \\
\addlinespace[0.3em]
& mHC$_{\mathrm{identity},n=2}$ & \qty{656}{\text{K}} 
& 26.23 & 0.59 & 23.28 & 49.62 & 5.42 & 25.69 
& \textcolor{myblue}{\underline{76.01}} 
& 25.69 \\
& mHC$_{\mathrm{identity},n=4}$ & \qty{2.36}{\text{M}} 
& 26.92 & 0.56 & 25.32 & 49.83 & 5.43 & 25.70 & 75.46 
& \textcolor{myblue}{\underline{27.74}} \\
& dynamic KromHC$_{n=4}$ & \qty{3.41}{\text{M}} 
& 22.47 & 1.64 & 22.67 & 49.66 & 2.16 & 24.45 & 75.41 & 15.34 \\
& mHC$_{\mathrm{identity},n=8}$ & \qty{8.91}{\text{M}} 
& 24.94 & 0.71 & 22.74 & 50.00 & 5.06 & 29.28 & 75.57 & 15.01 \\
\midrule
\midrule
\multirow{3}{*}{Combo}
& mHC$_{\mathrm{identity},n=2}$+VeRA & \qty{1.08}{\text{M}} 
& 26.08 & 1.45 & 27.29 
& \textcolor{myblue}{\underline{49.49}} 
& \textcolor{myblue}{\underline{3.18}} 
& \textcolor{myblue}{\underline{26.71}} 
& \textcolor{myblue}{\underline{75.41}} 
& \textcolor{myblue}{\underline{25.13}} \\
& static KromHC$_{n=16}$+LoRA$_{r=8}$ & \qty{6.03}{\text{M}} 
& 27.06 
& \textcolor{myblue}{\underline{4.01}} 
& 23.20 & 49.17 & 2.14 & 24.59 & 75.08 & 0.02 \\
& mHC$_{\mathrm{identity},n=2}$+LoRA$_{r=8}$ & \qty{6.68}{\text{M}} 
& \textcolor{myblue}{\underline{27.74}} 
& 2.73 
& \textcolor{myblue}{\underline{28.51}} 
& 49.37 & 0.82 & 24.38 & 74.86 & 0.62 \\
\bottomrule
\end{tabular}
}
\end{table*}

\textbf{Downstream benchmarks.}\quad
Table~\ref{tab:benchmark_results_1b} reports downstream performance on eight benchmark tasks for single methods and mHC+PEFT combinations.
The benchmark results are more task-dependent than the loss results.
No single method dominates on all tasks, and performance is not monotonic in trainable parameter count.
In light of efficiency-oriented evaluation, this means that adding parameters, increasing $n$, or combining adapters does not automatically improve downstream performance.
Within this mixed picture, the clearest pattern is again identity preservation: at equal $n$, mHC$_\mathrm{identity}$ beats learned residual mixing on most downstream tasks, consistent with the loss results in Table~\ref{tab:kromhc_expansion_identity}.
However, unlike the loss results, increasing $n$ does not uniformly improve downstream performance.

Since mHC adapts routing instead of weights, we also combine it with LoRA and VeRA at comparable budgets.
The outcome is split: combinations improve test loss, with, e.g., static mHC$_{n=16}$+LoRA$_{r=8}$ reaching \num{1.214} against \num{1.282} for LoRA$_{r=8}$ at just \num{1280} extra parameters, but they do not consistently beat the single methods on benchmarks.
Thus, the combination results are suggestive but not conclusive: adding mHC can help improve loss, but does not reliably translate to downstream gains.

\textbf{Scaling to 7B.}\quad
Finally, we test whether these findings transfer to a larger backbone.
We repeat selected runs on OLMo-2-7B, comparing mHC$_{\mathrm{identity}}$, LoRA at matched budgets, and their combination, see Appendix~\ref{ap:experiments_7B}.
The pattern matches 1B: mHC$_{\mathrm{identity}}$ improves over the frozen backbone, but standalone LoRA stays stronger.
Replacing one LoRA rank with mHC gives a small gain, \num{0.980} for mHC$_{\mathrm{identity},n=2}$+LoRA$_{r=13}$ versus \num{0.981} for LoRA$_{r=14}$, and downstream results remain mixed.
\section{Discussion \& Conclusion}\label{sec:discussion_conclusion}

This is the first study to apply mHC in finetuning rather than pre-training.
mHC lifts a frozen single-stream Transformer into an $n$-stream residual architecture, making residual topology adaptive without updating pre-trained weights.
As a standalone PEFT method, mHC does not beat established baselines such as LoRA.
Its contribution is instead architectural: mHC exposes residual topology as an adaptation axis and lets us ask which parts of that topology finetuning actually (re-)uses.

\textbf{Identity preservation.}\quad
mHC behaves differently in finetuning than how \citet{zhu2025hyperconnections} and \citet{xie2026mhcmanifoldconstrainedhyperconnections} introduced it for pre-training: $\mathbf{H}^{l}_{\mathrm{res}}$ drifts towards identity, i.e., $\mathbf{H}^{l}_{\mathrm{res}} \approx \mathbf{I}_n$.
Fixing $\mathbf{H}^{l}_{\mathrm{res}} = \mathbf{I}_n$ improves loss at every $n$ while using fewer parameters, suggesting that the useful capacity comes instead from the read/write gates, which control access to the residual streams.\footnote{The observed degradation of learning $\mathbf{H}^{l}_{\mathrm{res}}$ is not catastrophic forgetting \citep{kirkpatrick2017overcoming}: the backbone is frozen, pre-trained weights are not overwritten.
Instead, learning $\mathbf{H}^{l}_{\mathrm{res}}$ could interfere with how pre-trained representations propagate.}

\textbf{Adaptive routing.}\quad
Static and dynamic mHC differ in whether routing is fixed or input-dependent.
We find that dynamic variants perform better, suggesting that routing different inputs differently matters, though they use many more parameters and computation.
Among the dynamic variants, KromHC gives the best loss vs.\ parameter trade-off.
More broadly, mHC can be viewed as conditional routing over residual streams (loosely analogous to mixture-of-experts, but over residual streams instead of experts).
The benchmark results remain mixed, so we do not claim a conclusive gain over stronger PEFT baselines, but residual routing appears different from simply increasing LoRA rank.

\textbf{Efficiency costs.}\quad
Parameter efficiency of structural adapters should be interpreted cautiously, as we found parameter count is an incomplete measure.
For mHC, increasing $n$ does not change backbone computation or the key-value cache, but it does increase the residual state that must be stored and routed.
Unlike LoRA updates, this routing also cannot be merged into the backbone at inference.
As a result, increasing $n$ increases run time and energy more than increasing LoRA rank does.
Thus, the efficiency costs of mHC come from activation memory and throughput, not only from trainable parameters.
Reducing them will likely require fused routing kernels or sparser routing manifolds.

\textbf{Conclusion.}\quad
mHC shows that a pre-trained Transformer's residual topology is adaptable.
The useful finetuning mechanism, however, is not to relearn this topology, but to preserve it and learn read/write routing around it.
This makes residual routing a new architectural axis for efficient adaptation.
Our results are limited to OLMo-2 and one Tulu-based mixture, and our unfused PyTorch implementation likely overstates the final cost.
Future work should (\emph{i}) study when mHC usefully combines with existing PEFT methods and (\emph{ii}) test whether mHC yields different behaviour: better domain transfer, continual adaptation, interpretability, grounding, or controlled (re-)use of pre-trained representations.





\medskip

\bibliographystyle{plainnat}
\bibliography{references}

\clearpage
\appendix

\section{Efficiency beyond parameter count}\label{ap:efficiency}

mHC is parameter-efficient: static KromHC adapts a frozen 1B backbone with 384 trainable parameters (see Table~\ref{tab:mhc_variant_comparison}, Section~\ref{sec:experiments}), about 0.00002586\% of the backbone, and the dynamic identity-preserving variants also stay well under 1\% of the backbone parameters (see Table~\ref{tab:kromhc_expansion_identity}, Section~\ref{sec:experiments}).
But as we noted in Section~\ref{sec:experiments}, trainable parameter count does not capture the full cost of the method.
The read-in gate collapses the $n$ streams to a single $d$-vector before each frozen branch (see Eq.~\ref{eq:readin}), so backbone FLOPs and the key-value cache do not depend on $n$.
The residual state, however, is $(B, T, n, d)$ at every wrapped sub-layer (Figure~\ref{fig:wrapvisual}, Section~\ref{sec:method}) and also must be kept for the backward pass.
Freezing the backbone removes weight gradients and optimiser state, which is where PEFT usually saves memory (see, e.g., \citet{hu2022lora}).
It does not remove this activation cost though, and that cost grows with $n$ regardless of how few parameters are trained.

Also, fixing $\mathbf{H}^l_{\mathrm{res}} = \mathbf{I}_n$ does not resolve this.
It drops the residual projection and the mixing product but keeps the streams themselves, so the parameter saving becomes a smaller share of the routing budget as $n$ grows (Table \ref{tab:kromhc_expansion_identity}) while the activation cost keeps increasing. 

\begin{table}[h]
\caption{Training cost for LoRA and mHC$_{\text{identity}}$ on OLMo-2-7B,
trained for \num{20000} steps. Energy and power are GPU-only. Wall-clock time includes periodic validation passes. Only within-family comparisons are meaningful since they ran on different hardware types.}
\vspace{0.1em}
\label{tab:cost}
\centering
\footnotesize
\begin{tabular}{llrrrrr}
\toprule
Method & GPU & \# params & Time (h) & GPU energy (kWh) & Mean GPU power (W) \\
\midrule
LoRA$_{r=1}$   & A100 & 2.50\,M & 57.8 & 20.11 & 348  \\
LoRA$_{r=4}$   & A100 & 9.99\,M & 59.1 & 21.12 & 358  \\
LoRA$_{r=8}$   & A100 & 20.0\,M & 58.1 & 19.40 & 334  \\
LoRA$_{r=14}$  & A100 & 35.0\,M & 59.7 & 19.86 & 333  \\
\midrule
mHC$_{\text{identity},n=2}$ & H100 & 2.62\,M & 37.8 & 19.52 & 516 \\
mHC$_{\text{identity},n=4}$ & H100 & 9.44\,M & 48.8 & 23.66 & 486 \\
\bottomrule
\end{tabular}
\end{table}

The same pattern is visible in training cost, see Table~\ref{tab:cost} above.
For LoRA, changing the rank from $r=1$ to $r=14$ changes the number of trainable parameters by a factor of fourteen, but wall-clock time and GPU energy change only by a few \% and not monotonically, suggesting the low-rank updates are small compared with the frozen backbone computation.
For mHC, doubling the number of streams increases both time and energy by more than this variation.
Because GPU utilisation is similar and mean power is slightly lower, the extra cost likely comes from routing and memory movement rather than extra arithmetic.
At inference, the difference is structural: LoRA can be merged into $W_0$ (Eq.~\ref{eq:lora_update}, Section~\ref{ap:PEFT_baselines}) at no cost, whereas residual routing cannot.
So mHC is not inefficient; its costs sit in memory and throughput rather than in parameters, which makes it a useful case study for how efficiency in foundation models should be reported.
\section{Background: extended}\label{ap:background_full}

Residual connections \citep{he2016deep} add an identity path to each layer and are usually written as
\begin{equation}\label{eq:residual_connection}
\mathbf{z}^{l} = f^{l}(\mathbf{z}^{l-1}) + \mathbf{z}^{l-1},
\end{equation}
where $\mathbf{z}^{l-1}, \mathbf{z}^{l} \in \mathbb{R}^{d}$ are the input and output representations of layer $l$, and $f^{l}: \mathbb{R}^{d} \to \mathbb{R}^{d}$ is the learned transformation.
In Transformer blocks \citep{vaswani2017attention}, normalisation placement determines how direct this path is.
In \emph{post-norm}, layer normalisation (LN) is applied \emph{after} the residual addition
\begin{equation}\label{eq:post_norm}
    \mathbf{z}^{l}=\operatorname{LN}\!\left(\mathbf{z}^{l-1} + f^{l}(\mathbf{z}^{l-1})\right).
\end{equation}
\emph{Pre-norm blocks}, which preserve a more explicit residual stream, apply normalisation \emph{inside} the $f^{l}$ branch
\begin{equation}\label{eq:pre_norm}
    \mathbf{z}^{l}=\mathbf{z}^{l-1}+f^{l}\!\left(\operatorname{LN}(\mathbf{z}^{l-1})\right).
\end{equation}
This creates trade-off between pre-norm variants, where highly similar features in deeper layers can lead to representation collapse, and post-norm variants, where vanishing gradients may be reintroduced \citep{wang2026spannormreconcilingtrainingstability}.
This distinction matters for HC and mHC because both modify how the residual stream is routed over depth.

\subsection{Hyper-Connections}\label{sec:HC}

HC generalises the single residual stream in Eq.~\eqref{eq:residual_connection} by maintaining $n$ residual streams \citep{zhu2025hyperconnections}.
Let
\begin{equation}
    \mathbf{Z}^{l-1} =
    \begin{bmatrix}
    \mathbf{z}^{l-1}_{1} \\
    \vdots \\
    \mathbf{z}^{l-1}_{n}
    \end{bmatrix}
    \in \mathbb{R}^{n \times d}
\end{equation}
denote these $n$ streams before layer $l$, where each $\mathbf{z}^{l-1}_{i} \in \mathbb{R}^{d}$.
HC reads from the streams using $\mathbf{h}^{l}_{\mathrm{pre}} \in \mathbb{R}^{n}$:
\begin{equation}\label{eq:readin}
    \mathbf{x}^{l}=(\mathbf{h}^{l}_{\mathrm{pre}})^{\top}\mathbf{Z}^{l-1} \in \mathbb{R}^{d}.
\end{equation}
The layer output $f^{l}(\mathbf{x}^{l})$ is written back using $\mathbf{h}^{l}_{\mathrm{post}} \in \mathbb{R}^{n}$, while $\mathbf{H}^{l}_{\mathrm{res}} \in \mathbb{R}^{n \times n}$ mixes the residual streams:
\begin{equation}\label{eq:hc_update}
    \mathbf{Z}^{l}
    =
    \mathbf{H}^{l}_{\mathrm{res}}\mathbf{Z}^{l-1}
    +
    \mathbf{h}^{l}_{\mathrm{post}} \otimes f^{l}(\mathbf{x}^{l}),
\end{equation}
where $\otimes$ denotes the outer product that writes the layer output into each stream by $\mathbf{h}^{l}_{\mathrm{post}}$.
Here, $\mathbf{h}^{l}_{\mathrm{pre}}$, $\mathbf{H}^{l}_{\mathrm{res}}$, and $\mathbf{h}^{l}_{\mathrm{post}}$ are the effective routing objects.
In static HC, they are obtained from learned logits $\tilde{\mathbf{h}}^{l}_{\mathrm{pre}}$, $\tilde{\mathbf{H}}^{l}_{\mathrm{res}}$, and $\tilde{\mathbf{h}}^{l}_{\mathrm{post}}$.
Figure~\ref{fig:mlp_compare} shows schematically what these objects do, compared to an ordinary residual connection and a LoRA-wrapped one.

This construction lets the model learn how to route representations across residual streams and depths.
Moreover, HC is also \emph{extremely} efficient in parameters.
For expansion rate $n$, static HC learns $n^{2}+2n$ routing logits per layer: $n^{2}$ for $\tilde{\mathbf{H}}^{l}_{\mathrm{res}}$, and $n$ for both $\tilde{\mathbf{h}}^{l}_{\mathrm{pre}}$ and $\tilde{\mathbf{h}}^{l}_{\mathrm{post}}$.
The drawback is that the effective mixing matrix $\mathbf{H}^{l}_{\mathrm{res}}$ is not constrained to preserve the stability of the identity path.
When Eq.~\eqref{eq:hc_update} is unrolled over many layers, repeated products of these matrices amplify or suppress the residual streams, leading to unstable representations or gradients.

\begin{figure*}[h]
\centering
\resizebox{\textwidth}{!}{%
\begin{tikzpicture}[
    every node/.style={font=\Large},
    box/.style={
        rectangle, rounded corners=4pt,
        draw=black!40, line width=0.55pt,
        align=center
    },
    outerbox/.style={
        rectangle,
        draw=black!55,
        line width=0.75pt,
        rounded corners=5pt,
        minimum width=10.0cm,
        minimum height=5.5cm
    },
    frozenbox/.style={
        box,
        fill=gray!8,
        draw=gray!55,
        dashed,
        minimum width=2.9cm,
        minimum height=1.55cm
    },
    gatingbox/.style={
        box,
        fill=violet!12,
        draw=violet!50,
        minimum width=1.9cm,
        minimum height=0.62cm
    },
    routingbox/.style={
        box,
        fill=orange!15,
        draw=orange!50,
        minimum width=2.55cm,
        minimum height=2.95cm
    },
    sumcircle/.style={
        circle,
        draw=black!50,
        fill=white,
        minimum size=7mm,
        inner sep=0pt
    },
    arr/.style={
        -{Stealth[length=5pt,width=4pt]},
        line width=0.7pt,
        draw=black!65
    },
    lab/.style={
        font=\Large,
        text=black!70,
        align=center
    }
]

\node[outerbox] (midbox) at (0,0) {};
\node[frozenbox] (mlp2) at (0,-0.30) {MLP};
\node[sumcircle,  minimum size=4.8mm] (sum2) at (0,1.15) {$+$};

\draw[arr] ($(midbox.south)+(0,0.12)$) -- (mlp2.south);
\draw[] (mlp2.north) -- (sum2.south);
\draw[arr] (sum2.north) -- ($(midbox.north)+(0,-0.12)$);

\coordinate (midresstart) at ($(midbox.south)+(0,0.90)$);
\coordinate (midresleft)  at ($(midbox.west)+(1.15,0)$);

\draw[] (midresstart) -| (midresleft);
\draw[arr] (midresleft) |- (sum2.west);

\node[lab] at (0,-3.25) {MLP with residual};

\begin{scope}[shift={(11.5,6.4)}]

\node[
    outerbox,
    minimum width=10.0cm,
    minimum height=5.5cm
] (rightbox) at (0,-6.4) {};

\node[
    routingbox,
    minimum width=1.85cm,
    minimum height=1.75cm
] (hres) at (-2.05,-6.65)
    {$\mathbf{H}_{\mathrm{res}}$};

\node[
    frozenbox,
    minimum width=2.05cm,
    minimum height=1.05cm
] (mlp3) at (1.25,-6.88)
    {MLP};

\node[
    gatingbox,
    minimum width=1.70cm,
    minimum height=0.52cm
] (hpost) at (1.25,-5.94)
    {$\mathbf{h}_{\mathrm{post}}$};

\node[
    gatingbox,
    minimum width=1.70cm,
    minimum height=0.52cm
] (hpre) at (1.25,-7.82)
    {$\mathbf{h}_{\mathrm{pre}}$};

\node[sumcircle, minimum size=4.8mm] (sum31) at (1.00,-5.01) {$+$};
\node[sumcircle, minimum size=4.8mm] (sum32) at (1.59,-4.60) {$+$};

\coordinate (inL0) at (0.85,-9.03);
\coordinate (inR0) at (1.55,-9.03);

\coordinate (splitL) at (0.85,-8.62);
\coordinate (splitR) at (1.55,-8.42);

\draw[] (inL0) -- (splitL);
\draw[] (inR0) -- (splitR);

\coordinate (hpreInL) at ($(hpre.south west)!0.27!(hpre.south east)$);
\coordinate (hpreInR) at ($(hpre.south west)!0.67!(hpre.south east)$);

\draw[arr] (splitL) -- (hpreInL);
\draw[arr] (splitR) -- (hpreInR);


\coordinate (hresInL) at ($(hres.south)+(-0.25,0)$);
\coordinate (hresInR) at ($(hres.south)+( 0.25,0)$);

\draw[arr] (splitL) -| (hresInL);
\draw[arr] (splitR) -| (hresInR);

\coordinate (hresOutL) at ($(hres.north)+(-0.25,0)$);
\coordinate (hresOutR) at ($(hres.north)+( 0.25,0)$);

\coordinate (resTrackL) at ($(hresOutL)+(0,0.75)$);
\coordinate (resTrackR) at ($(hresOutR)+(0,1.15)$);

\draw[] (hresOutL) -- (resTrackL);
\draw[arr] (resTrackL) -- (sum31.west);

\draw[] (hresOutR) -- (resTrackR);
\draw[arr] (resTrackR) -- (sum32.west);

\coordinate (hpOutL) at ($(hpost.north west)!0.35!(hpost.north east)$);
\coordinate (hpOutR) at ($(hpost.north west)!0.70!(hpost.north east)$);

\draw[] (hpOutL) |- (sum31.south);
\draw[] (hpOutR) |- (sum32.south);

\coordinate (out1) at (1.00,-3.72);
\coordinate (out2) at (1.59,-3.72);

\draw[arr] (sum31.north) -- (out1);
\draw[arr] (sum32.north) -- (out2);

\node[lab] at (0,-9.65) {(m)HC-wrapped MLP};

\end{scope}
\begin{scope}[shift={(23.0,13.8)}]

\node[
    outerbox,
    minimum width=10.0cm,
    minimum height=5.5cm
] (lorabox) at (0,-13.8) {};

\node[
    frozenbox,
    minimum width=2.05cm,
    minimum height=1.05cm
] (loramlp) at (1.25,-13.95)
    {MLP};

\node[
    trapezium,
    trapezium left angle=75,
    trapezium right angle=75,
    draw=violet!50,
    fill=violet!12,
    line width=0.55pt,
    minimum width=1.55cm,
    minimum height=0.52cm,
    align=center
] (loraA) at (-1.30,-14.35)
    {$\mathbf{A}$};

\node[
    trapezium,
    trapezium left angle=105,
    trapezium right angle=105,
    draw=violet!50,
    fill=violet!12,
    line width=0.55pt,
    minimum width=1.55cm,
    minimum height=0.52cm,
    align=center
] (loraB) at (-1.30,-13.35)
    {$\mathbf{B}$};

\node[sumcircle, minimum size=4.8mm] (lorasum) at (1.25,-12.00) {$+$};

\coordinate (lorain0) at (1.25,-16.43);
\coordinate (lorasplit) at (1.25,-15.82);
\coordinate (loraStart) at ($(lorasplit)+(-2.55,0)$);

\draw[] (lorain0) -- (lorasplit);

\draw[] (lorasplit) -- (loraStart);

\draw[arr] (lorasplit) -- (loramlp.south);
\draw[] (loramlp.north) -- (lorasum.south);

\draw[arr] (loraStart) -| (loraA.south);


\coordinate (loraTrack) at ($(loraB.north)+(0,0.87)$);

\draw[] (loraB.north) -- (loraTrack);
\draw[arr] (loraTrack) -- (lorasum.west);

\coordinate (midresleft) at ($(lorabox.west)+(1.15,0)$);

\draw[] (lorasplit) -| (midresleft);
\draw[arr] (midresleft) |- (lorasum.west);

\coordinate (loraout) at (1.25,-11.12);

\draw[arr] (lorasum.north) -- (loraout);

\node[lab] at (0,-17.05) {LoRA-wrapped MLP};

\end{scope}
\end{tikzpicture}%
}
\caption{Comparison of an MLP with a residual connection, (m)HC, and a LoRA adapter.
}
\label{fig:mlp_compare}
\end{figure*}

\subsection{Manifold-Constrained Hyper-Connections}\label{sec:mHC}

mHC by \citet{xie2026mhcmanifoldconstrainedhyperconnections} addresses the instability of HC by constraining $\mathbf{H}^{l}_{\mathrm{res}}$.
The used constraint is doubly stochasticity: a matrix $\mathbf{H}^{l}_{\mathrm{res}} \in \mathbb{R}^{n \times n}$ is doubly stochastic \emph{iff} all its entries are nonnegative, i.e., $(\mathbf{H}^{l}_{\mathrm{res}})_{ij} \geq 0$, and all rows and columns sum to one, i.e., for all $i,j \in \{1,\ldots,n\}$,
\begin{equation}\label{eq:doubly_stochastic}
    \sum_{j=1}^{n}(\mathbf{H}^{l}_{\mathrm{res}})_{ij} = 1,
    \qquad
    \sum_{i=1}^{n}(\mathbf{H}^{l}_{\mathrm{res}})_{ij} = 1.
\end{equation}
\textbf{Sinkhorn normalisation.}\quad The set of doubly stochastic matrices is known as the Birkhoff polytope $\mathcal{B}_{n}$ \citep{birkhoff1946tres, yang2026mhclitedontneed20}.
mHC approximately projects the unconstrained residual mixing matrix onto this manifold using the Sinkhorn-Knopp algorithm \citep{sinkhorn1967concerning, xie2026mhcmanifoldconstrainedhyperconnections}:
\begin{equation}\label{eq:mhc_sinkhorn}
    \mathbf{H}^{l}_{\mathrm{res}}
    =
    \operatorname{Sinkhorn}\!\left(
        \tilde{\mathbf{H}}^{l}_{\mathrm{res}}
    \right),
\end{equation}
where $\tilde{\mathbf{H}}^{l}_{\mathrm{res}} \in \mathbb{R}^{n \times n}$.
Given logits $\tilde{\mathbf{H}} \in \mathbb{R}^{n \times n}$, we first form a positive matrix
\begin{equation}
    \mathbf{M}^{(0)} = \exp\left(\tilde{\mathbf{H}}\right).
\end{equation}
The Sinkhorn-Knopp procedure then alternates row and column normalisation as follows:
\begin{equation}
    \mathbf{M}^{(t + 1/2)}_{ij}
    =
    \frac{
        \mathbf{M}^{(t)}_{ij}
    }{
        \sum_{k=1}^{n}\mathbf{M}^{(t)}_{ik}
    },
    \qquad
    \mathbf{M}^{(t+1)}_{ij}
    =
    \frac{
        \mathbf{M}^{(t+1/2)}_{ij}
    }{
        \sum_{k=1}^{n}\mathbf{M}^{(t+1/2)}_{kj}
    }.
\end{equation}
For positive matrices, this procedure converges to a doubly stochastic matrix under standard conditions \citep{sinkhorn1967concerning}.
In practice, mHC uses a finite number of iterations.
Our implementation performs the normalisation in log space for numerical stability and uses 20 iterations, matching the original mHC configuration by \citet{xie2026mhcmanifoldconstrainedhyperconnections}.
Lastly, recent work by \citet{yang2026mhclitedontneed20} suggests that this iterative projection can be avoided through an exact doubly stochastic reparameterisation.

Through this normalisation, streams can still exchange information, but the router can no longer freely grow or shrink the residual signal across depth.
At the same time, the identity property of the original residual connection is preserved (which we verify in Appendix~\ref{ap:identity_property}).
Empirically, \citet{xie2026mhcmanifoldconstrainedhyperconnections} show that mHC keeps activations and gradients stable across layers, while unconstrained HC can amplify them substantially (see Figures \ref{fig:27b_forward_backward_gain} and \ref{fig:27b_loss_grad_all} in Appendix \ref{ap:results_manifold}).

\textbf{Static mHC.}\quad For the static variant used in our experiments, the remaining routing objects are obtained from learned logits as
\begin{equation}\label{eq:static_mhc_maps_main}
\begin{aligned}
    \mathbf{h}^{l}_{\mathrm{pre}}
    &=
    \sigma\!\left(
        \tilde{\mathbf{h}}^{l}_{\mathrm{pre}}
    \right),
    \\
    \mathbf{h}^{l}_{\mathrm{post}}
    &=
    2\sigma\!\left(
        \tilde{\mathbf{h}}^{l}_{\mathrm{post}}
    \right),
\end{aligned}
\end{equation}
where $\sigma(\cdot)$ is the element-wise sigmoid function and $\tilde{\mathbf{h}}^{l}_{\mathrm{pre}}, \tilde{\mathbf{h}}^{l}_{\mathrm{post}} \in \mathbb{R}^{n}$. 
Together, Eq.~\eqref{eq:mhc_sinkhorn} and Eq.~\eqref{eq:static_mhc_maps_main} give the constrained routing objects used in the HC update, Eq.~\eqref{eq:hc_update}.
The update, therefore, keeps the same read-in, write-out, and residual-mixing structure as HC, but now uses a doubly stochastic residual mixing matrix.
Since products of doubly stochastic matrices remain doubly stochastic, repeated routing over depth stays controlled.

\textbf{Dynamic mHC.}\quad
The parameterisation above is static: each layer learns one fixed set of routing logits that is used for all inputs.
\emph{Dynamic} HC and mHC instead make these routing logits input-dependent, where the model does not use the same $\tilde{\mathbf{h}}^{l}_{\mathrm{pre}}$, $\tilde{\mathbf{h}}^{l}_{\mathrm{post}}$, and $\tilde{\mathbf{H}}^{l}_{\mathrm{res}}$ for every example, but predicts them from the current residual streams \citep{zhu2025hyperconnections,xie2026mhcmanifoldconstrainedhyperconnections}.

Dynamic mHC first collects the streams into one flattened vector and then
normalises it with RMSNorm \citep{rmsnorm}:
\begin{equation}\label{eq:dynamic_mhc_flatten_normalisation}
\begin{aligned}
    \mathbf{u}^{l-1}
    &=
    \operatorname{vec}\!\left(\mathbf{Z}^{l-1}\right)
    \in \mathbb{R}^{nd},
    \\
    \bar{\mathbf{u}}^{l-1}
    &=
    \operatorname{RMSNorm}\!\left(\mathbf{u}^{l-1}\right).
\end{aligned}
\end{equation}
The normalised vector is then used to generate dynamic logits for the read-in,
write-out, and residual-mixing maps:
\begin{equation}\label{eq:dynamic_mhc_logits_main}
\begin{aligned}
    \tilde{\mathbf{h}}^{l}_{\mathrm{pre}}(\mathbf{Z}^{l-1})
    &=
    \alpha^{l}_{\mathrm{pre}}
    (\mathbf{\Phi}^{l}_{\mathrm{pre}})^{\top}
    \bar{\mathbf{u}}^{l-1}
    +
    \mathbf{b}^{l}_{\mathrm{pre}},
    \\
    \tilde{\mathbf{h}}^{l}_{\mathrm{post}}(\mathbf{Z}^{l-1})
    &=
    \alpha^{l}_{\mathrm{post}}
    (\mathbf{\Phi}^{l}_{\mathrm{post}})^{\top}
    \bar{\mathbf{u}}^{l-1}
    +
    \mathbf{b}^{l}_{\mathrm{post}},
    \\
    \tilde{\mathbf{r}}^{l}_{\mathrm{res}}(\mathbf{Z}^{l-1})
    &=
    \alpha^{l}_{\mathrm{res}}
    (\mathbf{\Phi}^{l}_{\mathrm{res}})^{\top}
    \bar{\mathbf{u}}^{l-1}
    +
    \mathbf{b}^{l}_{\mathrm{res}},
\end{aligned}
\end{equation}
where $\mathbf{\Phi}^{l}_{\mathrm{pre}},\mathbf{\Phi}^{l}_{\mathrm{post}}\in\mathbb{R}^{nd\times n}$, $\mathbf{b}^{l}_{\mathrm{pre}},\mathbf{b}^{l}_{\mathrm{post}}\in\mathbb{R}^{n}$, and $\mathbf{\Phi}^{l}_{\mathrm{res}}\in\mathbb{R}^{nd\times n^{2}}$, $\mathbf{b}^{l}_{\mathrm{res}}\in\mathbb{R}^{n^{2}}$.
The scalars $\alpha^{l}_{\mathrm{pre}},\alpha^{l}_{\mathrm{post}},\alpha^{l}_{\mathrm{res}}$ control the strength of the dynamic component.
The residual logits are reshaped into a matrix
\begin{equation}\label{eq:dynamic_mhc_res_matrix_main}
    \tilde{\mathbf{H}}^{l}_{\mathrm{res}}(\mathbf{Z}^{l-1})
    =
    \operatorname{mat}\!\left(
        \tilde{\mathbf{r}}^{l}_{\mathrm{res}}(\mathbf{Z}^{l-1})
    \right)
    \in\mathbb{R}^{n\times n}.
\end{equation}
The constrained routing objects are then obtained with the same maps as in the
static Sinkhorn variant:
\begin{equation}\label{eq:dynamic_mhc_constrained_maps_main}
\begin{aligned}
    \mathbf{h}^{l}_{\mathrm{pre}}(\mathbf{Z}^{l-1})
    &=
    \sigma\!\left(
        \tilde{\mathbf{h}}^{l}_{\mathrm{pre}}(\mathbf{Z}^{l-1})
    \right),
    \\
    \mathbf{h}^{l}_{\mathrm{post}}(\mathbf{Z}^{l-1})
    &=
    2\sigma\!\left(
        \tilde{\mathbf{h}}^{l}_{\mathrm{post}}(\mathbf{Z}^{l-1})
    \right),
    \\
    \mathbf{H}^{l}_{\mathrm{res}}(\mathbf{Z}^{l-1})
    &=
    \operatorname{Sinkhorn}\!\left(
        \tilde{\mathbf{H}}^{l}_{\mathrm{res}}(\mathbf{Z}^{l-1})
    \right).
\end{aligned}
\end{equation}
Dynamic Sinkhorn mHC thus keeps the doubly stochastic residual constraint, but lets the constrained router depend on the current representation, increasing expressivity.

This increased expressivity comes at the cost of more parameters.
For one layer, it uses
\begin{equation}\label{eq:dynamic_mhc_param_count_main}
    nd(n^{2}+2n) + (n^{2}+2n) + 3
\end{equation}
routing parameters: projection weights, biases, and three learned dynamic gates.
In contrast, the static Sinkhorn parameterisation uses only $n^{2}+2n$
routing logits per layer.
Dynamic mHC is thus more expressive, but no longer has the same extremely
small parameter budget as static mHC.

\subsection{mHC variants: mHC-lite \& KromHC}

The Sinkhorn parameterisation is the most expensive part of mHC: it starts from $n^{2}$ logits and obtains $\mathbf{H}^{l}_{\mathrm{res}}\in\mathcal{B}_{n}$ by iterative row and column normalisation.
Several recent variants (e.g., \citealp{wang2026acceleratingbirkhoffprojectionmanifoldconstrained, 11476522, lyubinin2026tbpmhcexpressivitymanifoldconstrainedhyper, dandachi2026gomhcdirectparameterizationmanifoldconstrained, sengupta2026jpmhcdynamicalisometryorthogonal, liu2026birkhoffpolytopespectralsphereconstrainedhyperconnections}) keep the same read-in and write-out structure, but replace the way $\mathbf{H}^{l}_{\mathrm{res}}$ is parameterised.
We focus on mHC-lite by \citet{yang2026mhclitedontneed20} and KromHC by \citet{zhou2026kromhcmanifoldconstrainedhyperconnectionskroneckerproduct} because they provide two direct and complementary alternatives to the Sinkhorn parameterisation of $\mathbf{H}^{l}_{\mathrm{res}}$: mHC-lite keeps the full Birkhoff polytope but removes iterative normalisation, whereas KromHC instead restricts the residual mixer to matrices built from small doubly stochastic factors.

\textbf{mHC-lite.}\quad
mHC-lite removes the Sinkhorn projection by using the Birkhoff-von Neumann theorem \citep{yang2026mhclitedontneed20}.
Since every doubly stochastic matrix can be written as a convex combination of permutation matrices, it parameterises the residual mixer as
\begin{equation}
    \mathbf{H}^{l}_{\mathrm{res}}
    =
    \sum_{k=1}^{n!}
    a^{l}_{k}
    \mathbf{P}_{k},
\end{equation}
where $\{\mathbf{P}_{k}\}_{k=1}^{n!}$ are the $n!$ permutation matrices, and $\mathbf{a}^{l} = \operatorname{softmax}(\boldsymbol{\theta}^{l})$ is a distribution over permutation matrices.
This gives an exactly doubly stochastic matrix without iterative normalisation.
The dynamic version uses the same conditioning mechanism as dynamic Sinkhorn mHC, but generates the $n!$ permutation logits instead of the $n^{2}$ Sinkhorn logits.
The drawback is that the number of permutation matrices is $n!$, so mHC-lite is simple and exact for small $n$, but does not scale well to many streams.
  
\textbf{KromHC.}\quad
KromHC addresses the scalability problem from a different direction \citep{zhou2026kromhcmanifoldconstrainedhyperconnectionskroneckerproduct}.
Instead of representing the full Birkhoff polytope, it constructs $\mathbf{H}^{l}_\mathrm{res}$ from a Kronecker product of smaller doubly stochastic matrices
\begin{equation}
    \mathbf{H}^{l}_{\mathrm{res}}
    =
    \mathbf{A}^{l}_{1}
    \otimes
    \mathbf{A}^{l}_{2}
    \otimes
    \cdots
    \otimes
    \mathbf{A}^{l}_{\log_{2} n},
\end{equation}
where $\mathbf{A}^{l}_{k}\in\mathcal{B}_{2}$ and $n$ is a power of $2$.
Because the Kronecker product of doubly stochastic matrices is again doubly stochastic, KromHC preserves the constraints of Sinkhorn mHC.
The difference is that it covers just a subset of matrices that can be written in this Kronecker form.
This makes KromHC less expressive but much cheaper to parameterise.

In the dynamic version, the same conditioning mechanism as dynamic Sinkhorn mHC generates logits for the small factors $\mathbf{A}^{l}_{k}$ instead of for the full $\mathbf{H}^{l}_\mathrm{res}$ explicitly, which is cheaper.
In pre-training experiments, KromHC matches or outperforms prior mHC variants while using fewer trainable parameters \citep{zhou2026kromhcmanifoldconstrainedhyperconnectionskroneckerproduct}. A drawback of this method is that $n$ must be a power of two.

\section{Methods: extended}\label{ap:method_full}

Here we provide additional detail to the methods of Section~\ref{sec:method}.
Our implementation is available on GitHub: \url{https://github.com/valentijn7/mHC_PEFT.git}.

\textbf{Backbone.}\quad
We use OLMo-2-0425-1B as the main pre-trained backbone.
OLMo-2 is a family of open LMs with public checkpoints, code, logs, and training details \citep{olmo20252olmo2furious}.\footnote{Although our experiments focus on LMs, the construction only assumes residual blocks and is not specific to text data.}
The model is large enough to meaningfully test PEFT behaviour, while small enough to (\emph{i}) run controlled experiments over multiple baselines, (\emph{ii}) do multiple ablations, and (\emph{iii}) run benchmarks, all within computational budget.
To test whether findings persist at a larger scale while limiting compute, we evaluate selected adapters on the corresponding OLMo-2-1124-7B checkpoint.\footnote{Both the 1B- and 7B-model are downloadable through Hugging Face, via \url{https://huggingface.co/allenai/OLMo-2-0425-1B} and \url{https://huggingface.co/allenai/OLMo-2-1124-7B}, respectively.}
Unless stated otherwise, all original backbone parameters are frozen and only the PEFT method parameters at hand are updated.

\subsection{Static mHC as PEFT}\label{sec:method_smhc}

We adapt the pre-trained Transformer by wrapping each attention and MLP sub-layer with a static mHC module, i.e., where the routing logits $\tilde{\mathbf{h}}^{l}_{\mathrm{pre}}$, $\tilde{\mathbf{H}}^{l}_{\mathrm{res}}$, and $\tilde{\mathbf{h}}^{l}_{\mathrm{post}}$ are learned per layer and kept fixed at inference, and with routing objects $\mathbf{h}^{l}_{\mathrm{pre}}$, $\mathbf{H}^{l}_{\mathrm{res}}$, and $\mathbf{h}^{l}_{\mathrm{post}}$.
For the Sinkhorn variant, this wrapper uses the static parameterisation from Section~\ref{sec:mHC}.
For mHC-lite and KromHC, we start from the public dynamic implementations and adapt them to the same static PEFT setting by removing input-dependent conditioning.\footnote{Public codebase for mHC-lite: \url{https://github.com/FFTYYY/mhc-lite}; and for KromHC: \url{https://github.com/wz1119/KromHC}.}

The wrapper preserves the original attention and MLP computations: the wrapped branch still receives one hidden representation and returns one hidden representation.
OLMo does not use the traditional post-norm of Eq.~\eqref{eq:post_norm}; instead, normalisation is applied only to the sub-layer output before it is added to the residual stream
\begin{equation}\label{eq:olmo_residual_update}
    \mathbf{z}^{l}
    =
    \mathbf{z}^{l-1}
    +
    \operatorname{LN}\!\left(
        f^{l}(\mathbf{z}^{l-1})
    \right).
\end{equation}
Thus, the residual stream itself is not normalised directly, and stacked sub-layers modify it through additive normalised updates.
Internally, however, this representation is lifted to $n$ residual streams.
After the final layer, the streams are averaged to recover the standard hidden representation.
Thus, the module changes only the residual routing around the frozen branches, while preserving the input-output interface of the pre-trained Transformer.
This makes the wrapper compatible with Hugging Face-style Transformer modules without changing the surrounding architecture; see Figure~\ref{fig:wrapvisual} for a schematic overview.

\textbf{Trainable parameters.}\quad
For static Sinkhorn mHC, each wrapped sub-layer trains only the routing logits
$\tilde{\mathbf{h}}^{l}_{\mathrm{pre}}$,
$\tilde{\mathbf{h}}^{l}_{\mathrm{post}}$, and
$\tilde{\mathbf{H}}^{l}_{\mathrm{res}}$,
giving, for $n=4$, $n^2+2n = 24$ trainable parameters per wrapped sub-layer.
Since OLMo-2-1B has $16$ Transformer layers and we wrap both attention and MLP branches and since the softmax readout adds $n$ parameters, this gives $16 \cdot 2 \cdot 24 + 4 = 772$ trainable mHC parameters in total.
The counts for mHC-lite and KromHC differ only in the number of residual-mixing logits used to parameterise $\mathbf{H}^{l}_{\mathrm{res}}$; for $n=4$, this gives $n!+2n=32$ parameters per sub-layer for mHC-lite and $2\log_{2}(n)+2n=12$ for KromHC. The parameter count does not capture the full cost (App. \ref{ap:efficiency}). The read-in gate collapses the $n$ streams before each frozen branch (Eq. \ref{eq:readin}), so backbone FLOPs and the KV cache do not depend on $n$. The residual state, however, gains a stream dimension at every wrapped sub-layer and must be kept for the backward pass. Freezing the backbone removes weight gradients and optimiser state, but not this activation cost. Fixing $\mathbf{H}^l_{\mathrm{res}} = \mathbf{I}_n$ removes routing parameters while leaving the streams themselves.

\subsection{Dynamic mHC as PEFT}\label{sec:dynamic_mHC_as_PEFT}

We instantiate dynamic mHC as a PEFT module.
The wrapper is inserted in the same locations as in the static setting, but the routing objects are no longer fixed after training, as described in Section~\ref{sec:mHC}.
For the Sinkhorn variant, we implement the dynamic formulation of \citet{xie2026mhcmanifoldconstrainedhyperconnections} directly in PyTorch.
For mHC-lite and KromHC, we start from their public dynamic implementations and adapt them to the same PEFT interface.
To make the dynamic variants comparable, we use the same wrapped branches, stream expansion, readout convention, and diagnostics.

\subsection{Initialisation}\label{ap:identity_property}
All variants introduce a stream permutation symmetry: all $n$ residual streams start as identical copies of the same hidden representation, and symmetric routing can make their gradients identical as well.
We therefore initialise the module to stay close to the original residual behaviour of the pre-trained model, adding small perturbations that let the streams differentiate during training.

\textbf{Identity property.}\quad
The mHC module should behave like a standard residual connection at initialisation, because pre-trained transformer weights are trained expecting residual connections. An initial behaviour similar to standard residual connections has several benefits. Primarily, it ensures that finetuning starts from OLMo's optimised pre-training maximum. This prevents using the capacity of mHC to claw back to OLMo-equivalent behaviour from a degraded starting state. Secondly, it follows the identity-mapping principle of residual connections. The added capacity is initially inactive but becomes active during training. It is activated gradually, rather than forcing the network to recover from a disrupted state. Lastly, when mHC is identical to vanilla OLMo at initialisation, any improvement or degradation that is subsequently measured is attributable purely to what mHC learns during training, not to a coincidentally favourable or unfavourable initialisation. 

\textbf{Initialisation of routing objects.}\quad The routing objects in each mHC module are initialised to preserve this property. $\mathbf{H}^{l}_{\mathrm{res}}$ should be a doubly stochastic matrix to preserve the mean residual signal, which is done by setting $\tilde{\mathbf{H}}^{l}_{\mathrm{res}}$ logits to zero and applying Sinkhorn, giving $\mathbf{H}^{l}_{\mathrm{res}} = \frac1n \mathbf{11}^T$. $\mathbf{h}^{l}_{\mathrm{pre}}$ is initialised as the uniform average $\frac{1}{n}\mathbf{1}$. $\tilde{\mathbf{h}}^{l}_{\mathrm{post}}$ logits are set to zero, making $\mathbf{h}^{l}_{\mathrm{post}} = 2\sigma(0) = 1$, thereby writing back the stream output to every stream with weight one. This property was tested by checking whether the vanilla OLMo and mHC-wrapped OLMo model's final logits are sufficiently numerically close. Further, we directly test the residual identity equation by comparing the output of one mHC-wrapped sub-layer against $x + \mathrm{branch}(x)$. If this holds, the mHC wrapper behaves like a standard residual connection at initialisation.

\textbf{Symmetry breaking.}\quad\label{ap:symmetry_breaking}
At initialisation all $n$ streams are identical copies of the same hidden representation.
If the routing objects are symmetric as well, the gradients are symmetric too, so the streams stay identical and the model remains at an $S_n$ stream-permutation fixed point.
The purpose of multiple streams is then defeated, as the streams never diverge and specialise.
The conditions of residual equivalency conditions constrain how this symmetry can be broken.
A doubly stochastic $\mathbf{H}^{l}_{\mathrm{res}}$ maps identical streams back to the same streams, and $\mathbf{h}^{l}_{\mathrm{post}} = \mathbf{1}$ writes the branch output into every stream with equal weight, so neither can distinguish the streams without also breaking residual equivalence.
Breaking it must thus be done with the read-in gate $\mathbf{h}^{l}_{\mathrm{pre}}$.

\textbf{One-hot read-in.}\quad
In all variants except static Sinkhorn mHC, we break the symmetry deterministically and per layer.
Sub-layer $l$ selects stream $$k = l \bmod n,$$ and gives it logit $8$, while the remaining logits are set s.t.\ $\sum_{s=1}^{n} h^{l}_{s,\mathrm{pre}} = 1$ holds exactly.
Hence, like this, the branch input still equals $x$ while the streams are identical, but only the selected stream receives gradient from the read path and the selection differs across layers, which lets the streams differentiate.
At the same time, $\mathbf{h}^{l}_{\mathrm{post}}$ keeps a zero bias, i.e.\ $\mathbf{h}^{l}_{\mathrm{post}} = \mathbf{1}$, and $\mathbf{H}^{l}_{\mathrm{res}}$ starts at the identity.

Dynamic Sinkhorn mHC realises this with a diagonal bias of $8$ on the residual logits, KromHC and mHC-lite by putting all mass on the identity permutation and a logit of $-12$ on the remaining ones.
In the dynamic variants the projections are zero-initialised and the gates $\alpha^{l}_{\mathrm{pre}}$, $\alpha^{l}_{\mathrm{post}}$, and $\alpha^{l}_{\mathrm{res}}$ start at $10^{-2}$, so routing starts at its bias value and only becomes input-dependent during training.
For the initialisation ablation of Appendix~\ref{app:experiment_initialisation_residual}, $\mathbf{H}^{l}_{\mathrm{res}}$ can instead start uniform, at a non-identity permutation, or at the identity with bounded off-diagonal mass, where the leaked mass is capped at $0.9$ so the selected permutation stays the dominant one.

\textbf{Static Sinkhorn mHC.}\quad
Static Sinkhorn mHC breaks the symmetry with noise rather than with a one-hot read-in.
We add Gaussian noise ($\sigma = 10^{-4}$) to the read-in, write-out, and residual logits, and renormalise $\mathbf{h}^{l}_{\mathrm{pre}}$ afterwards to maintain the convex combination property $\sum_{s=1}^{n} h^{l}_{s,\mathrm{pre}} = 1$, which is what makes the branch input equal the embedding when the streams are identical at layer $0$.
The noise on $\mathbf{H}^{l}_{\mathrm{res}}$ needs no such correction, because Sinkhorn returns a doubly stochastic matrix for any logits.
Its residual logits start at zero, so $\mathbf{H}^{l}_{\mathrm{res}} = \frac{1}{n}\mathbf{1}\mathbf{1}^{\top}$ at initialisation rather than the identity used by the other variants.

\textbf{Residual logit dropout.}\quad
The static Sinkhorn module further supports Bernoulli dropout on the residual logits, where random entries are set to $-\infty$ before Sinkhorn normalisation, with a check that prevents fully masked rows or columns.
This targets the matrix where cross-stream exchange happens, while Sinkhorn's row- and column-stochasticity guarantees no magnitude distortion on the residual stream.
We keep it disabled in all reported runs, as the initialisation already breaks the symmetry, and dropout is never applied at inference.

\textbf{Residual equivalence vs.\ symmetry-breaking trade-off.}\quad Altough the symmetry-breaking perturbations technically break exact residual-equivalence between the mHC-wrapped model and vanilla OLMo, this is not a concern in practice. Before each run, we verify empirically that the resulting deviation stays within a tight numerical tolerance. The final logits of the wrapped and vanilla models remain close, and the single-sub-layer output stays within roughly $1\%$ of $x+\mathrm{branch}(x)$. A deviation of this magnitude is negligible relative to the scale of the hidden states (and is comparable to the rounding error already introduced by \texttt{bf16} precision). The model thus still starts from a point that is, for all practical purposes, OLMo's pre-trained minimum, i.e.\ the loss at initialisation is effectively unchanged, gradients remain small, and no early instability is introduced.

\subsection{Training data and preprocessing}\label{sec:method_data}

\textbf{Dataset.}\quad
We finetune on the Tulu-3 SFT OLMo mixture dataset \citep{lambert2024tulu}, an instruction-response dataset from 18 sources designed for OLMo and not reported as part of its pre-training \citep{olmo20252olmo2furious}.
It is downloadable through Hugging Face.
It covers tasks such as dialogue, question answering, and reasoning.

\textbf{Stratified split.}\quad
Because Tulu-3 combines multiple source datasets, we sample examples proportionally to their source labels and then create a fixed train/validation/test split shared by all methods.
The main split contains \num{320000} training, \num{2500} validation, and \num{2500} test samples.
The procedure first counts the number of examples belonging to each source in the full dataset. For a target subset size $n$, it then assigns each source a sampling budget proportional to its original frequency: $n_s = \lfloor n \cdot \frac{N_s}{N} \rfloor$, where $N_s$ is the number of examples from source $s$, and $N$ is the total dataset size. The remaining samples are distributed to the sources with the largest fractional remainders. Finally, examples are randomly sampled without replacement from each source using a fixed seed, and the selected indices are shuffled. This produces a reduced dataset with approximately the same source distribution as the original dataset.

\textbf{Preprocessing.}\quad
We tokenize the data with the OLMo-2-0425-1B tokenizer, use a maximum sequence length of 2048 tokens, and pack tokenized examples into fixed-length blocks to reduce padding.
The packed splits are cached and reused for all methods.
Sequence-length statistics and batching details for GPU optimisation are given in Appendix~\ref{ap:sequence_length}.

\subsection{Baselines and training setup}\label{sec:method_training}

\textbf{Training setup.}\quad
We train all methods with next-token cross-entropy loss and keep the backbone frozen unless stated otherwise.
All main runs use \num{20000} optimisation steps, with only method-specific PEFT parameters varying across methods.
We report hyperparameter in Appendix~\ref{ap:hyperparameters}.
Before each full training, we verify routing constraints, gradient flow, and frozen-backbone equivalence.

\textbf{Carbon emissions.}\quad
Most experiments are run on single NVIDIA A100 Tensor Core GPUs of the Dutch national supercomputer, Snellius.
The total carbon emissions for this study is approximately 99.45 $\mathrm{kgCO}_2\mathrm{eq}$. Emissions were estimated using CodeCarbon \citep{codecarbon}; the estimated carbon intensity of the Netherlands was 143~$\mathrm{g\,CO_2eq\,kWh^{-1}}$ in May~2026 \citep{nowtricity2026co2}. 

\textbf{Diagnostics.}\quad
Before launching full training runs, we run a set of diagnostics on each wrapped model.
These checks verify that the model preserves the expected causal-LM output shape, that the routing modules expose valid $\mathbf{h}^{l}_{\mathrm{pre}}$, $\mathbf{H}^{l}_{\mathrm{res}}$, and $\mathbf{h}^{l}_{\mathrm{post}}$ objects, and that $\mathbf{H}^{l}_{\mathrm{res}}$ is finite, nonnegative, square, and doubly stochastic up to numerical tolerance.
We also check that gradients flow to the intended trainable parameters, that frozen backbone parameters receive no gradients, and that the wrapped model is numerically equivalent to the frozen backbone at initialisation.
Finally, we run a short synthetic overfitting test to verify that the residual streams can diverge once training starts.

\subsection{Evaluation}\label{sec:method_evaluation}

\textbf{Loss and perplexity.}\quad
During training, we monitor training loss. Validation scores are used for monitoring and hyperparameter choices. Performance is measured on the validation splits using cross-entropy loss and perplexity,
\begin{equation}
    \operatorname{PPL} = \exp(\mathcal{L}),
\end{equation}
where $\mathcal{L}$ is the average next-token cross-entropy.
\clearpage
\section{Experiments: extended}\label{ap:experiments}

This appendix supports the identity-preservation finding with variant comparisons
(Section~\ref{ap:comparing_mHC_variants}), routing analyses
(Section~\ref{sec:main_interpret}), ablations
(Sections~\ref{sec:ablation_H_res} to \ref{ap:weight_experiments}), and learning rate tuning
(Section~\ref{app:lr_tuning}).

\subsection{Comparing mHC variants}\label{ap:comparing_mHC_variants}
 
We first compare the three parameterisations: Sinkhorn mHC, mHC-lite, and KromHC.
For each, we evaluate both the static and dynamic version with expansion rate $n=4$, trained for \num{3000} steps.
Table~\ref{tab:mhc_variant_comparison} shows that all mHC variants improve over the frozen OLMo-2-1B baseline while training only a small number of parameters.
The dynamic variants consistently outperform their static counterparts, which is expected because of their greater model capacity.
Dynamic KromHC obtains the lowest train and test loss among the tested variants.
Also, we can report near-identical stream-divergence trajectories in the first 50 training steps, consistent with the theoretical expectation that all three preserve the doubly-stochastic residual-mixing property.

\textbf{Parameter efficiency.}\quad
The number of trainable parameters for each method can be found in Table~\ref{tab:mhc_variant_comparison}. The static variants are relatively tiny. 
Static KromHC has the lowest number of parameters and trains only \num{384} parameters, corresponding to $0.00002586\%$ of the OLMo-2-1B backbone.
Despite this, it reduces test loss from \num{2.00} to \num{1.77}, matching static mHC-lite and slightly outperforming static Sinkhorn mHC.
Similarily, dynamic KromHC outperforms the other mHC parametrisation and trains approximately \qty{3.41}{\text{M}} parameters, corresponding to $0.22\%$ of the backbone.

\textbf{Choice of default.}\quad
Among the dynamic variants, KromHC is both the most parameter-efficient and the performs the best too.
This suggests that the Kronecker restriction does not hurt performance in this setting, while it does reduce the routing parameter count.
We therefore use dynamic KromHC as the default mHC parameterisation in the remaining experiments, unless stated otherwise.

We also investigated the learning rate in Appendix~\ref{app:lr_tuning}:
a value of \texttt{1e-3} gives the best and most stable performance, and is used in the main dynamic KromHC experiments.

\subsection{Interpretability}\label{sec:main_interpret}

To gain a deeper understanding of mHC as a PEFT method, we conduct interpretability experiments.

Recent work on (m)HC showed that residual mixing often remains close to identity in pre-training and that signal and interpretable features concentrate in a dominant stream \citep{alimaskina2026analyzingstreamcollapsehyperconnections}. A comparable dominance pattern was observed at the causal level, where certain streams exerted significant influence on model output while representationally similar streams remained largely passive \citep{peng2026ablaterescuecausalanalysis}.

\begin{wrapfigure}{r}{0.48\linewidth}
  \vspace{-1.4em}
  \centering
  \includegraphics[width=0.98\linewidth]{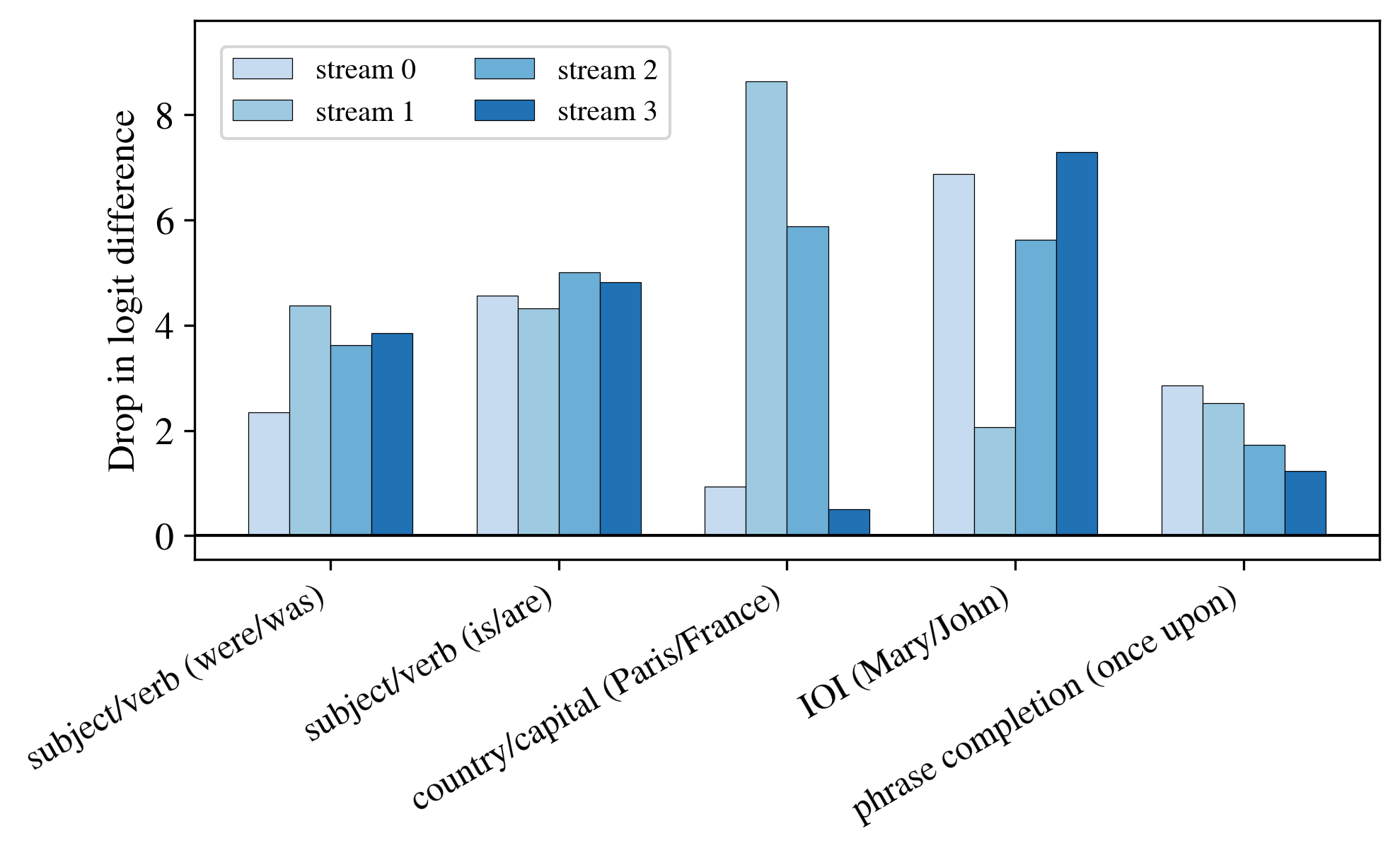}
  \caption{For each prompt, we zero one stream's routing matrices and measure the change in the correct-incorrect logit margin. Taller bars indicate stronger causal sensitivity to that stream, thus revealing prompt-dependent stream roles.}
  \label{fig:stream_ablation}
  \vspace{-1.0em}
\end{wrapfigure}

To examine whether our method shows the same patterns, we apply two methods.
With per-stream logit lens we investigate the confidence and differences in the prediction over layers per stream. Moreover, since representational similarity between streams has been shown to not necessarily reflect their functional or causal role \citep{peng2026ablaterescuecausalanalysis}, we complement logit lens and weight visualisation with stream ablations.
Stream ablation causally verifies the effect of each stream for different language tasks (subject/verb, country/capital, IOI, phrase completion). Also, we examine the routing over layers by applying linear centered kernel alignment (CKA).
Accordingly, we visualise $\mathbf{H}^{l}_{\mathrm{res}}$ for selected layer pairs, focusing on layers flagged by the CKA analysis.

\begin{figure}[h]
    \centering
    \begin{minipage}[c]{0.45\linewidth}
        \centering
        \includegraphics[height=7.0cm, keepaspectratio]{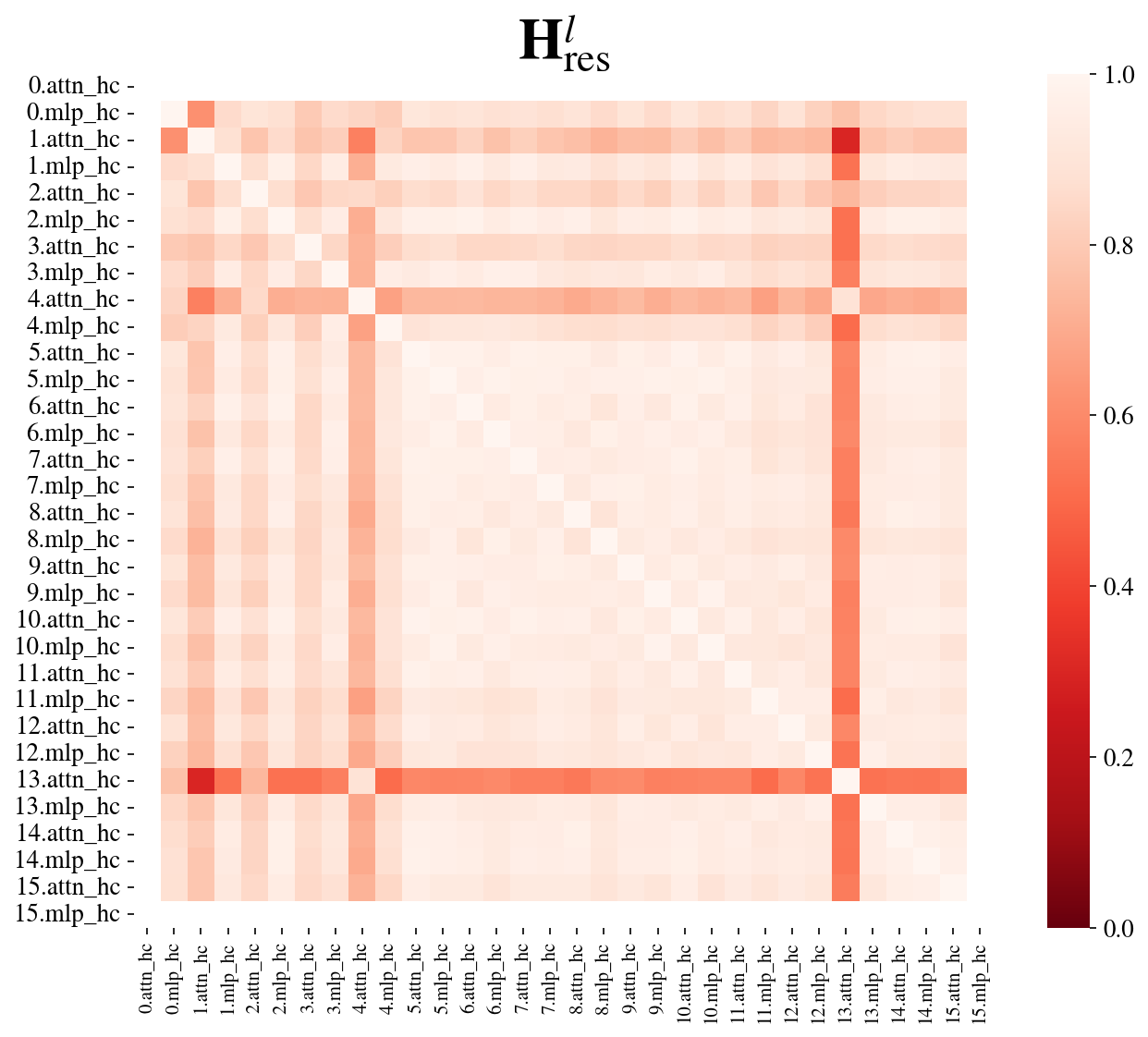}
    \end{minipage}
    \hfill
    \begin{minipage}[c]{0.49\linewidth}
        \centering
        \includegraphics[width=\linewidth, height=3.35cm, keepaspectratio]{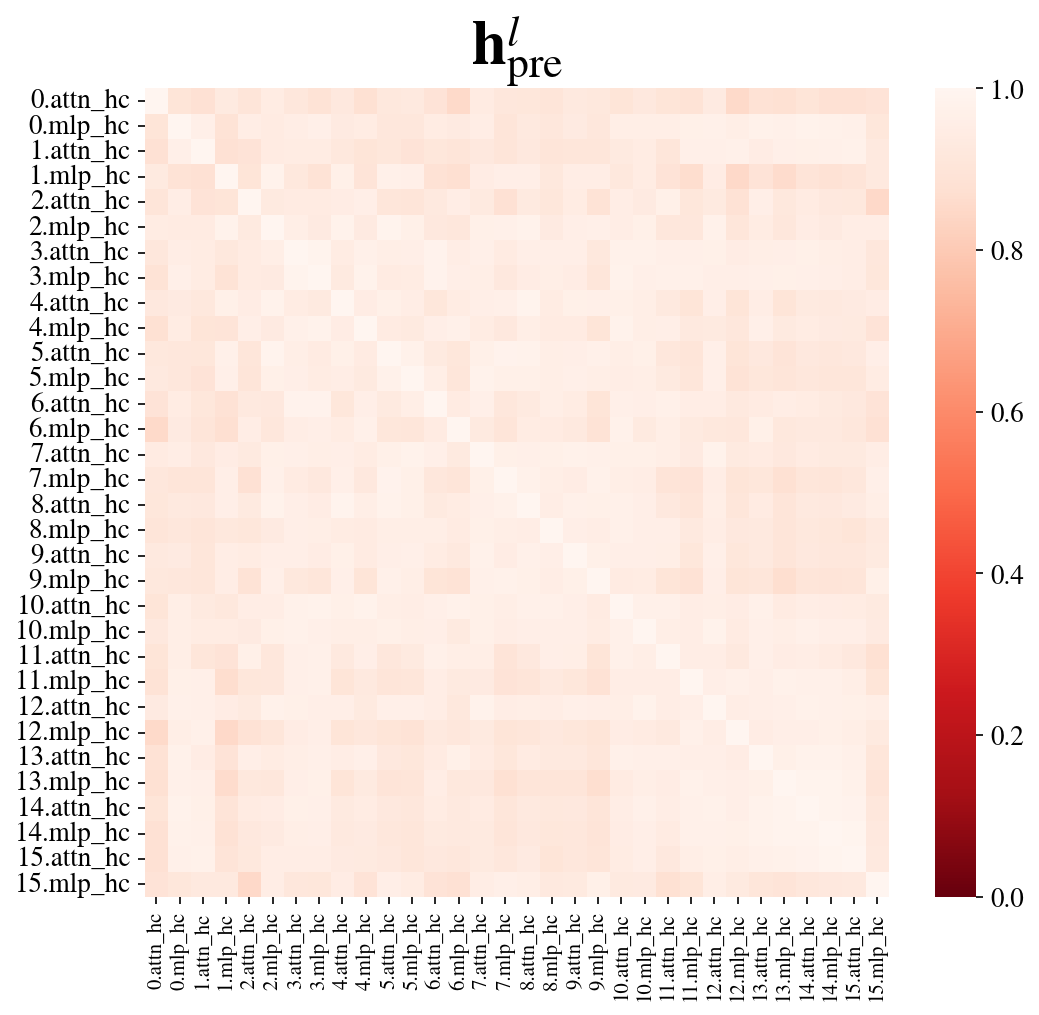}\\[0.3cm]
        \includegraphics[width=\linewidth, height=3.35cm, keepaspectratio]{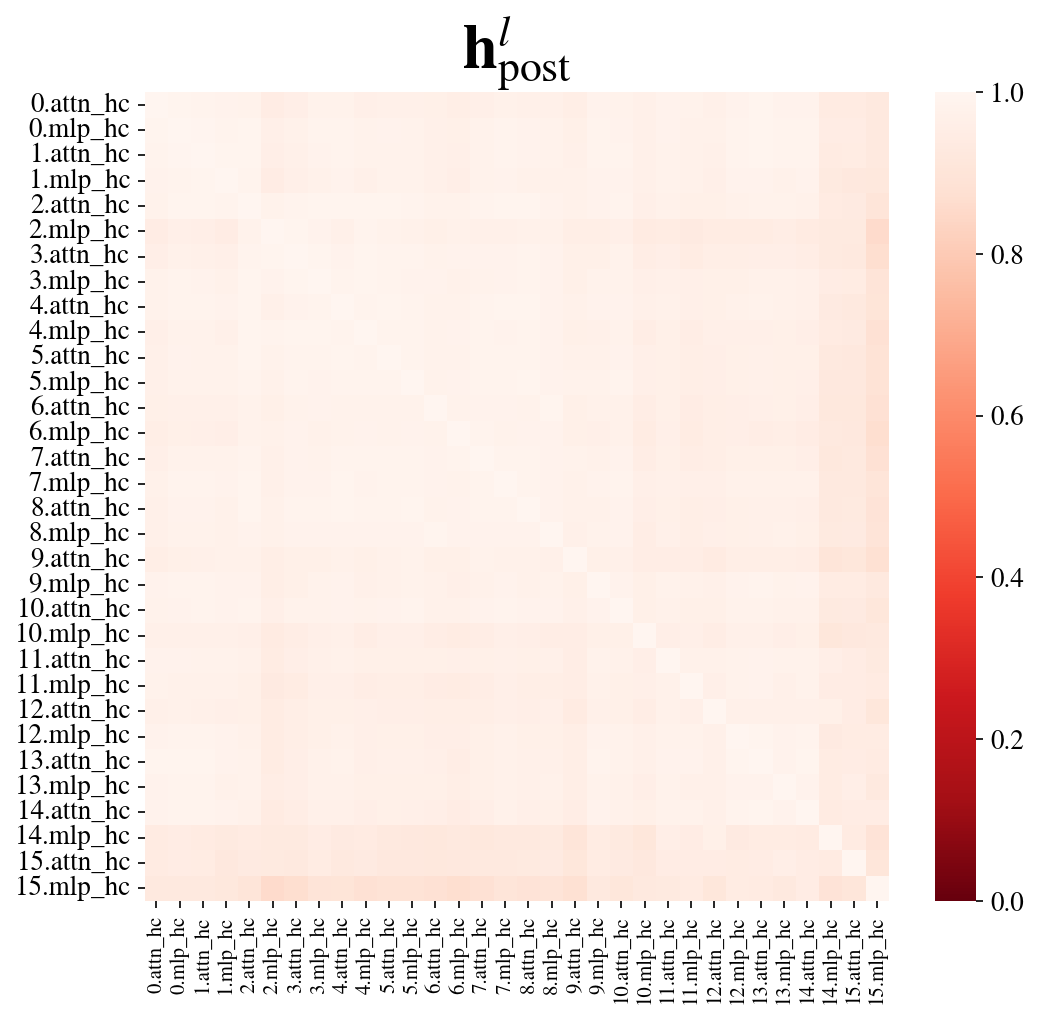}
    \end{minipage}
    \caption{The heatmap shows the linear centered kernel alignment (CKA) between all layers.}
    \label{fig:cka_layers}
\end{figure}

\begin{figure}[h]
  \centering
  \includegraphics[width=0.9\linewidth]{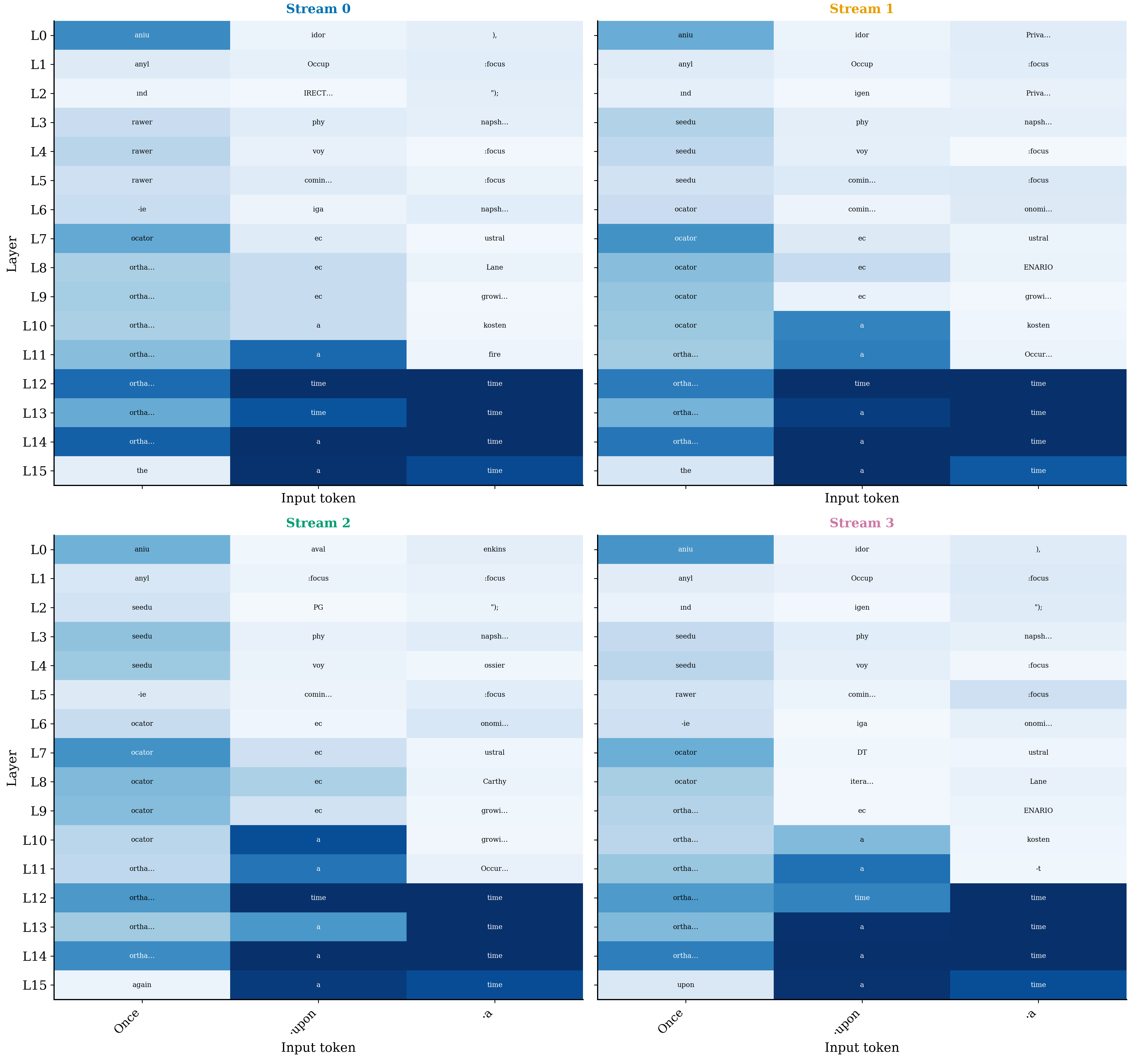}
  \includegraphics[width=0.9\linewidth]{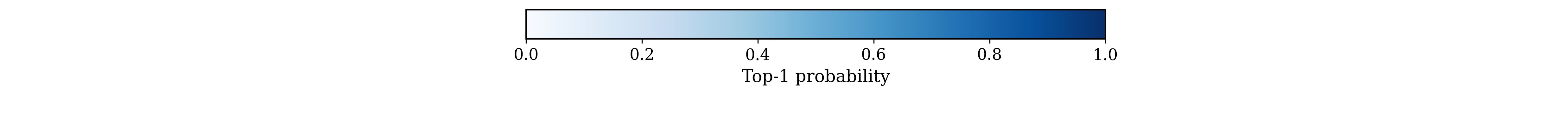}
  \caption{Each panel shows the top-1 predicted token at each layer and position for one KromHC stream. For high-confidence completions, KromHC routing does not induce stream specialisation.}
    \label{fig:logit_per_stream}
\end{figure}

\clearpage

\begin{wrapfigure}{r}{0.48\linewidth}
  \vspace{-1.0em}
  \centering
  \includegraphics[width=0.98\linewidth]{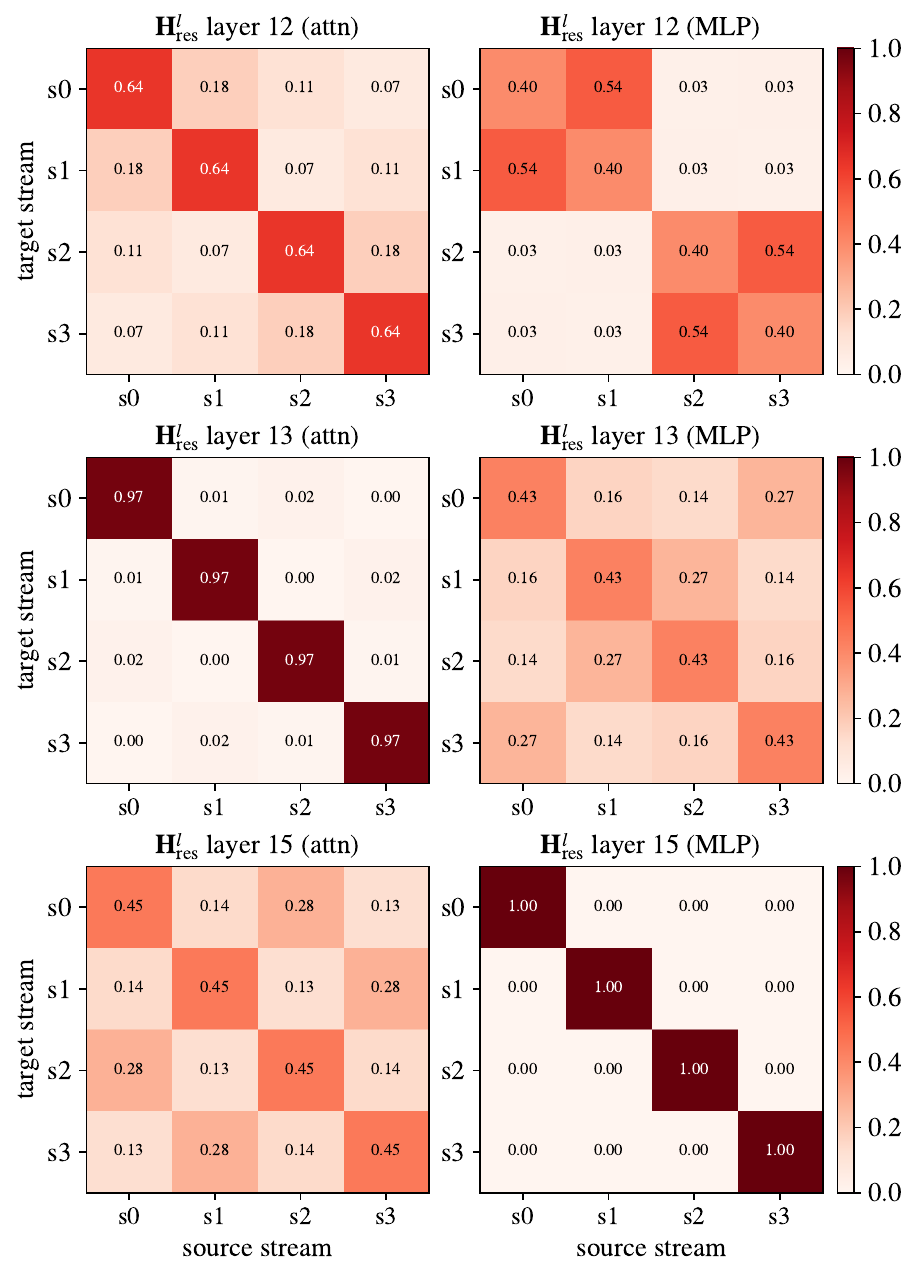}
  \caption{Visualisation of \(\mathbf{H}^{l}_{\mathrm{res}}\) in static mHC for attention and MLP sub-layers in layers 12, 13, and 15. After earlier layers start mixing routes, layer 13 returns to a strong identity matrix.}
  \label{fig:interpretability_layer_12_13_15}
  \vspace{-1.0em}
\end{wrapfigure}

Figures \ref{fig:cka_layers}, \ref{fig:stream_ablation}, and \ref{fig:logit_per_stream} shows the results.
First, routing is found to differ by layer, where most differences can be found in layer 4 and 13.
After visualising the weights in these interesting layers from CKA (see Figure \ref{fig:interpretability_layer_12_13_15}), we can see that later layers (specifically layer 13) show a shift back to near-identity.
Moreover, some layers mix across all streams while others almost entirely do not so.
The CKA findings highlight that pre-layer hidden representations are more differentiated across layers, whereas the post-layer representations are substantially more similar. Based on the observed CKA patterns, we hypothesise that $\mathbf{h}^{l}_{\mathrm{pre}}$ might play a more significant role, as it exhibit greater representational diversity than the corresponding $\mathbf{h}^{l}_{\mathrm{post}}$. Secondly, regarding stream specialisation, there is limited evidence of a dominant stream. The stream specialisation patterns are indicated to be  prompt-dependent, with its level varying by task. However, since prior work did not focus upon task or prompt types, it remains unclear whether this prompt-dependence is specific to finetuning or a pattern that has simply not yet been examined in pre-training. 

Most strikingly, $\mathbf{H}^{l}_{\mathrm{res}}$ was found to shift back to near-identity in later layers, of which Figure~\ref{fig:interpretability_layer_12_13_15} contains an example.  
This pattern is, for instance, visible in layer 13 attention and layer 15 MLP, where the diagonal entries dominate and the off-diagonal entries are close to zero, indicating that the residual streams are propagated almost independently. The finding is in line with earlier interpretability research of mHC for pre-training, where deeper mixing was also found to close to identity in mHC-lite \citep{alimaskina2026analyzingstreamcollapsehyperconnections}.
It suggests that the model learns to suppress cross-stream mixing through $\mathbf{H}^{l}_{\mathrm{res}}$ rather than exploit it (despite breaking symmetry initially). At the same time, this suppression is not uniform, as other layers show some extent of cross-stream mixing (e.g., layer 12). One hypothesis could be that $\mathbf{h}^{l}_{\mathrm{pre}}$ and $\mathbf{h}^{l}_{\mathrm{post}}$ play a more essential role than $\mathbf{H}^{l}_{\mathrm{res}}$ itself, which we discuss next.

\subsection{Ablation: residual mixing components}\label{sec:ablation_H_res}

To further investigate the interpretability findings and to isolate the contribution of each of the three routing objects ($\mathbf{h}^{l}_{\mathrm{pre}}$, $\mathbf{H}^{l}_{\mathrm{res}}$, and $\mathbf{h}^{l}_{\mathrm{post}}$) and their pairwise interactions, we run ablation experiments.
Each routing object in mHC has a distinct role: $\mathbf{h}^{l}_{\mathrm{pre}}$ selects the branch input from the streams,  $\mathbf{H}^{l}_{\mathrm{res}}$ mixes streams across depth, and $\mathbf{h}^{l}_{\mathrm{post}}$ writes back to the streams.
Ablations are implemented by replacing a component with its identity-preserving neutral form, consistent with the work by \citet{xie2026mhcmanifoldconstrainedhyperconnections} (while making sure the diagnostics of Section~\ref{sec:method_training} still pass).
$\mathbf{h}^{l}_{\mathrm{pre}}$ is replaced by uniform weights of $1/n$, $\mathbf{H}^{l}_{\mathrm{res}}$ by the identity matrix, and $\mathbf{h}^{l}_{\mathrm{post}}$ by uniform ones.
For efficiency, we adapt the code of KromHC such that ablating $\mathbf{H}^{l}_{\mathrm{res}}$ additionally drops its residual coefficients entirely, so that component's parameters are removed rather than merely fixed. 

We perform six runs with $n=4$ streams and train for $3000$ steps.
We individually ablate each object and the possible pairs.
Table \ref{tab:mhc_ablation} reports the results.
The findings indicate that $\mathbf{h}^{l}_{\mathrm{pre}}$ and $\mathbf{h}^{l}_{\mathrm{post}}$ are crucial for performance. This is in line with the suggestions of the interpretability experiments (discussed in Section \ref{sec:main_interpret}).   

Notably, these results are in contrast with the findings of \citet{xie2026mhcmanifoldconstrainedhyperconnections}, where the residual mapping was linked to higher performance gains in pre-training.
We hypothesise that this can be linked to a fundamental discrepancy between pre-training and finetuning. 
In pre-training, representations are learned from scratch and $\mathbf{H}^{l}_{\mathrm{res}}$ can effectively use stream routing. 
In contrast, when finetuning, the base model already has rich representation in its residual streams. 
By imposing to mix residual streams at every layer with normalisation, pre-trained representation may be disrupted. 
Potentially, keeping $\mathbf{H}^{l}_{\mathrm{res}}$ as an identity matrix might be key in the use of mHC for finetuning.

\begin{table}[H]
\centering
\caption{Ablation of dynamic KromHC components, trained for \num{3000} steps. A check indicates the component is active. }
\vspace{0.1em}
\label{tab:mhc_ablation}
\setlength{\tabcolsep}{4.5pt}
\footnotesize
\begin{tabular}{ccc|ccc}
\toprule
$\mathbf{H}^{l}_{\mathrm{res}}$ & $\mathbf{h}^{l}_{\mathrm{pre}}$ & $\mathbf{h}^{l}_{\mathrm{post}}$ & $\mathcal{L}_\mathrm{train}$ & \textbf{$\Delta\mathcal{L}_\mathrm{train}$ vs mHC} & $\mathcal{L}_\mathrm{test}$ \\
\midrule
 \textcolor{mygreen}{  \checkmark  }         &       \textcolor{mygreen}{  \checkmark }       &      \textcolor{mygreen}{ \checkmark   }       & 1.38  & ---    &   1.32     \\
\textcolor{myred}{ \xmark }  &       \textcolor{mygreen}{ \checkmark }       &      \textcolor{mygreen}{ \checkmark }         & 1.43 & $+$0.05 & 1.36 \\   \textcolor{mygreen}{ \checkmark }
             & \textcolor{myred}{ \xmark } &    \textcolor{mygreen}{ \checkmark  }          & 1.52       & $+$0.14 & 1.46   \\  \textcolor{mygreen}{ \checkmark }
             &        \textcolor{mygreen}{ \checkmark}        & \textcolor{myred}{ \xmark } & 1.50    & $+$0.12 & 1.43   \\
             \textcolor{myred}{ \xmark }  & \textcolor{myred}{  \xmark}  &        \textcolor{mygreen}{ \checkmark    }    & 1.52       & $+$0.14 & 1.46   \\
               \textcolor{myred}{  \xmark} & \textcolor{mygreen}{ \checkmark    }   &        \textcolor{myred}{  \xmark}   & 1.50      & $+$0.12 & 1.43   \\
              \textcolor{mygreen}{ \checkmark    }  & \textcolor{myred}{  \xmark}  &        \textcolor{myred}{  \xmark}   & 2.02  & $+$0.64 & 2.00        \\

\bottomrule
\end{tabular}
\end{table}

\subsection{Ablation: initialisation of $\mathbf{H}^{l}_{\mathrm{res}}$}\label{app:experiment_initialisation_residual} 

\citet{zhu2025hyperconnections} initialise $\mathbf{H}^{l}_{\mathrm{res}}$ as an identity matrix, while \citet{xie2026mhcmanifoldconstrainedhyperconnections} do not specify how it is initialised.
We therefore set $\mathbf{H}^{l}_{\mathrm{res}} = \mathbf{I}_n$ at the start, up to numerical precision of \texttt{1e-4} due to noise added for symmetry breaking.
We observed in some initial tests that $\mathbf{H}^{l}_{\mathrm{res}}$ initially diverges from the identity matrix, but quickly moves back to it. 
This raised the question whether identity is genuinely a preferred solution to model returns to, or merely a local optimum induced by initialisation. 

As discussed in Section~\ref{ap:identity_property}, we initialise mHC so that the wrapped model starts from the same function as the original OLMo model.
Because each tested $\mathbf{H}^{l}_{\mathrm{res}}$ is row-stochastic, it maps identical streams back to the same streams.
Together with unit write-out weights, this means that the wrapped model is initially equivalent to the frozen backbone.

We test whether the choice of $\mathbf{H}^{l}_{\mathrm{res}}$ initialisation matters after training.
For $n=4$, we compare four initialisations for \num{3000} steps: identity, uniform, permutation, and noise, where uniform initialisation uses a uniform doubly stochastic matrix, permutation initialisation swaps streams, and noise initialisation starts from the identity and adds bounded off-diagonal mass.
We find that the choice of initialisation has little effect on the final result.
For all four, training converges to similar losses, and the learned residual mixers move back toward an identity-like structure.
This supports the interpretation that, in this finetuning setting, learned cross-stream residual mixing is not the main source of performance.
Instead, the model tends to preserve the pre-trained residual pathway even when $\mathbf{H}^{l}_{\mathrm{res}}$ is initialised away from the identity.

\subsection{Ablation: readout}\label{ap:weight_experiments}

During the experiments we found that the final MLP layer was completely uniform, as in Figure \ref{fig:interpretability_layer_12_13_15}, caused by the mean pool at the end of the static Sinkhorn mHC module.
To test whether this was of any influence, we tested substituting this mean readout with a softmax.
Surprisingly, we found that after after \num{3000} steps, the final $\mathbf{H}^{l}_{\mathrm{res}}$ was not uniform anymore, though when we continued training to \num{20000} steps, the matrix returned back to uniform.
Accordingly, after \num{20000} we found little difference in performance.

\subsection{Learning rate tuning}\label{app:lr_tuning}

We performed hyperparameter tuning to obtain a suitable learning rate (LR) for mHC.
Too high an LR can destabilise the warm-start by pushing the dynamic routing terms away from their identity-equivalent initialisation, whereas too low an LR can leave them near it.
All experiments use a cosine LR schedule with fixed $3\%$ warmup.
We sweep LR~$\in\{\texttt{1e-3}, \texttt{5e-4}, \texttt{1e-4}, \texttt{5e-5} \}$ and train dynamic KromHC with $n=4$ for $3000$ steps.
We cap the sweep at \texttt{1e-3} deliberately.
Beyond preserving consistency with the high end of typical PEFT learning rates, larger rates risk perturbing the identity-equivalent initialisation before the model has adapted, undermining the warm-start property that motivates the manifold constraint.
We therefore treat this sweep as selecting a \emph{stable} operating point for the subsequent experiments rather than as a search for a globally optimal LR. 

We observe an upward trend when increasing the LR from \texttt{5e-5} to \texttt{1e-3}, though the increase progressively diminishes. An
LR of \texttt{1e-3} achieved the best test loss while retaining stable warm-started training, and this LR is carried into the longer main experiments.

\subsection{mHC combined with other PEFT methods}\label{sec:experiments_mHC_combined}

We next test whether mHC can complement existing PEFT methods.
This is motivated by the fact that mHC modifies residual routing instead of local weight transformations like LoRA en VeRA. It thus may provide benefits that are complementary to methods such as LoRA and VeRA. We evaluate mHC$_{\mathrm{identity},n=2}$+VeRA, static mHC$_{n=16}$+LoRA$_{r=8}$, and mHC$_{\mathrm{identity},n=2}$+LoRA$_{r=8}$.
Standalone methods are included as references, and LoRA$_{r=9}$ is included as an approximate parameter-matched comparison for mHC$_{\mathrm{identity},n=2}$+LoRA$_{r=8}$.

\begin{figure}[b]
    \vspace{-1em}
    \centering
    \includegraphics[width=0.45\textwidth]{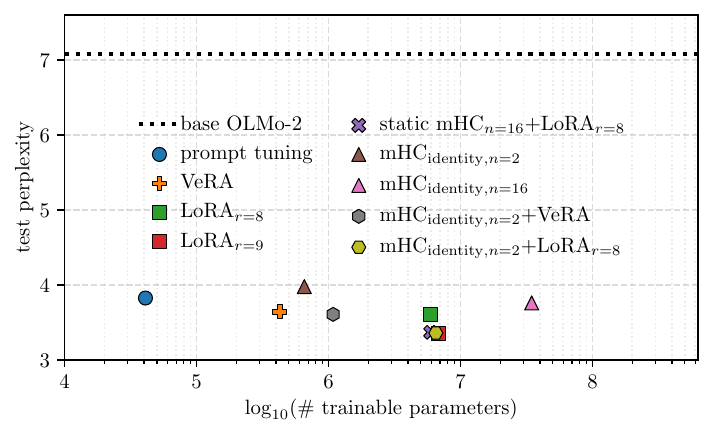}\vspace{-1em}
    \caption{Test perplexity (PPL) vs.\ \# trainable parameters.}
    \label{fig:test_ppl_vs_trainable_params}
\end{figure}

Figure~\ref{fig:test_ppl_vs_trainable_params} shows that combining mHC with LoRA can improve the trade-off between perplexity and the number of parameters. The combinations yield improved performance for only a small increase in the number of trainable parameters. In particular, static mHC$_{n=16}$+LoRA$_{r=8}$ reduces test loss compared with LoRA$_{r=8}$ with \num{1.282} to \num{1.214}, with only 1280 more parameters.

The results on benchmarks can be found in Table~\ref{tab:benchmark_results_1b} in Section~\ref{sec:experiments}.
In contrast to the improvement for test perplexity for combinations, the findings across downstream tasks are more mixed. For example, mHC$_{\mathrm{identity},n=2}$+LoRA$_{r=8}$ improves over LoRA$_{r=8}$ on BBH, MMLU, PIQA, and TriviaQA, but is worse on DROP, GSM8K, HellaSwag, and MATH.
Similarly, static mHC$_{n=16}$+LoRA$_{r=8}$ improves over LoRA$_{r=8}$ on BBH, MMLU, and PIQA, but not on most others. Overall, none of the combinations consistently outperform their standalone methods on all benchmarks. This suggests that the interaction between residual routing and existing PEFT methods might be task-dependent. So, mHC could be useful as a complementary routing mechanism, but benefits and/or drawbacks depends on the task. This finding that mHC and PEFT combinations are not consistently better at this scale motivates the larger-scale comparison in Section~\ref{ap:experiments_7B}.

\subsection{mHC with larger models}\label{ap:experiments_7B}

\begin{table}[h]
\vspace{-0.5em}
\centering
\caption{
OLMo-2-7B results for mHC, LoRA, and their combination.
All are trained for \num{20000} steps.
}
\label{tab:olmo7b_results}
\setlength{\tabcolsep}{3.5pt}
\footnotesize
\begin{tabular}{lccc}
\toprule
Method & \# params & $\mathcal{L}_\mathrm{train}$ & $\mathcal{L}_\mathrm{test}$ \\
\midrule
OLMo-2-7B & \num{0} & -- & 1.770 \\
\midrule
mHC$_{\mathrm{identity}, n=2}$ & \qty{2.62}{\text{M}} & 1.143 & 1.095 \\
LoRA$_{r=1}$ & \qty{2.50}{\text{M}} & 1.053 & 1.016 \\
\midrule
mHC$_{\mathrm{identity}, n=4}$ & \qty{9.44}{\text{M}} & 1.122 & 1.076 \\
LoRA$_{r=4}$ & \qty{9.99}{\text{M}} & 1.023 & 0.995 \\
\midrule
LoRA$_{r=14}$ & \qty{35.0}{\text{M}} & 1.004 & 0.981 \\
mHC$_{\mathrm{identity}, n=2}$+LoRA$_{r=13}$ & \qty{35.1}{\text{M}} & 1.001 & 0.980 \\
\bottomrule
\end{tabular}
\end{table}

We train selected adapters on OLMo-2-1124-7B to test whether the main findings transfer to a larger model.
We compare mHC$_\mathrm{identity}$ to LoRA at approximately matched parameter budgets: for mHC the budget is controlled by expansion rate $n$, for LoRA by rank $r$ \citep{hu2022lora}.
Matching the parameters isolates both inductive biases and directly compares low-rank weight updates with multi-stream residual routing.
All are trained for \num{20000} steps.

Table~\ref{tab:olmo7b_results} shows that the trends for OLMo-2-1B found in previous sections also hold at larger scale.
For example, mHC$_{\mathrm{identity}}$ substantially improves over the frozen OLMo-2-7B baseline in both training and test loss.
However, at approximately matched parameter budgets, LoRA achieves a lower train and test loss as a standalone PEFT method: LoRA$_{r=1}$ outperforms mHC$_{\mathrm{identity}, n=2}$, and LoRA$_{r=4}$ outperforms mHC$_{\mathrm{identity}, n=4}$.
This suggests that low-rank weight updates are more effective than residual-stream routing when each is used alone.

The combination result is more promising.
Replacing one LoRA rank with mHC parameters improves loss minimally: mHC$_{\mathrm{identity}, n=2}$+LoRA$_{r=13}$ gets a $\mathcal{L}_\mathrm{test}$ of \num{0.980} vs.\ \num{0.981} for LoRA$_{r=14}$ at nearly equal parameter count.
The gain is small, so the result should be interpreted cautiously.
Still, it indicates that residual routing does not simply duplicate the effect of LoRA; it is consistent with the idea that mHC and LoRA affect different model parts: LoRA changes local weight transformations and mHC changes how representations are routed through residual streams.

Table~\ref{tab:benchmark_results_7b} gives a more balanced picture than the loss results alone and shows downstream benchmark results.
At nearly matched parameter budget, mHC$_{\mathrm{identity},n=2}$+LoRA$_{r=13}$ improves over LoRA$_{r=14}$ on 4 out of 8 benchmarks: GSM8K, HellaSwag, PIQA, and TriviaQA, while LoRA is stronger on the others.
These results suggest that sometimes one LoRA rank can be replaced by an mHC routing module and give task-dependent improvement on downstream tasks.

\begin{table*}[t]
\centering
\caption{Downstream benchmark results for OLMo-2-7B, at matched parameter
budget, trained for \num{20000} steps. Metrics are EM for BBH, GSM8K, MATH and
TriviaQA; F1 for DROP; normalised accuracy for HellaSwag; accuracy for MMLU and
PIQA. Higher is better. Best scores
\textcolor{myblue}{\underline{underlined}}.}
\label{tab:benchmark_results_7b}
\setlength{\tabcolsep}{4pt}
\resizebox{\textwidth}{!}{
\begin{tabular}{llcccccccc}
\toprule
Method & \# trainable params & BBH & DROP & GSM8K & HellaSwag & MATH & MMLU & PIQA & TriviaQA \\
& & 3-shot & 3-shot & 8-shot & 10-shot & 4-shot & 5-shot & 0-shot & 5-shot \\
\midrule
LoRA$_{r=14}$ & \qty{35.0}{\text{M}} & \textcolor{myblue}{\underline{47.57}} & \textcolor{myblue}{\underline{6.08}} & 58.53 & 60.00 & \textcolor{myblue}{\underline{5.48}} & \textcolor{myblue}{\underline{59.75}} & 79.27 & 0.03 \\
mHC$_{\mathrm{id},n=2}$+LoRA$_{r=13}$ & \qty{35.1}{\text{M}} & 45.80 & 3.93 & \textcolor{myblue}{\underline{60.65}} & \textcolor{myblue}{\underline{60.30}} & 3.68 & 57.06 & \textcolor{myblue}{\underline{79.60}} & \textcolor{myblue}{\underline{0.07}} \\
\bottomrule
\end{tabular}
}
\end{table*}
\clearpage
\section{Hyperparameters}\label{ap:hyperparameters}

This appendix gives more detail on the hyperparameter configurations and choices.

\subsection{High-level GPU memory optimisation}\label{ap:sequence_length}

For determining the sequence length where we strike a balance between speed and data usage, we derive Tulu statistics, see Table~\ref{tab:tulu_length_stats}.
For the experiments, we used a maximum sequence length of 2048 tokens.
This threshold ensures that we cover more than 99\% of examples in \texttt{allenai/tulu-3-sft-olmo-2-mixture}. We applied the same size to our other experiments.

\begin{table}[h]
\centering
\caption{
Tokenized sequence length statistics for the T\"ulu 3 OLMo SFT mixture.
Lengths are computed after applying the OLMo chat formatting and tokenizing with the
\texttt{allenai/OLMo-2-1B} tokenizer. Fractions are the proportion of examples
going over each truncation threshold. 
}
\label{tab:tulu_length_stats}
\footnotesize
\begin{tabular}{llr}
\toprule
Type & Statistic & Value \\
\midrule
Dataset & Hugging Face ID & \texttt{allenai/tulu-3-sft-olmo-2-mixture} \\
Tokenizer & Hugging Face ID & \texttt{allenai/OLMo-2-1B} \\
\midrule
Token length & $p_{50}$ & 262 \\
Token length & $p_{75}$ & 470 \\
Token length & $p_{90}$ & 741 \\
Token length & $p_{95}$ & 953 \\
Token length & $p_{99}$ & \num{1643} \\
Token length & max & \num{16440} \\
\midrule
Truncation rate & $>1024$ tokens & 4.045\% \\
Truncation rate & $>2048$ tokens & 0.540\% \\
Truncation rate & $>4096$ tokens & 0.085\% \\
\bottomrule
\end{tabular}
\end{table}

For optimising the effective batch size for the finetuning set-up, we tested multiple batch sizes on a single A100 GPU using OLMo-2-0425-1B with LoRA.
To get a conservative memory estimate, we fixed the maximum sequence length at 2048 tokens and padded examples to this length during the test.
We then increased the batch size while adjusting gradient accumulation to keep the effective batch size constant.

A per-device batch size of 4 with 4 gradient accumulation steps was the largest stable configuration, giving an effective batch size of 16.
Larger sizes, such as 8, resulted in CUDA out-of-memory errors under the worst-case (i.e.\ with padding to 2048 tokens) setting.
Thus, we use the 4-4 configuration for all main finetuning runs.
For some smaller experiments, we adjusted the batch size of 8 and 2 gradient accumulation steps.

\subsection{Hyperparameters of PEFT methods}\label{ap:PEFT_hyperparameters}

All experiments use a Hydra-managed configuration file.
The default values are stored in \texttt{config.yaml}; individual runs override only the method, parameter budget, training length, or learning rate when stated explicitly.
Table~\ref{tab:global_hyperparameters} lists the shared training setup, while Tables~\ref{tab:peft_hyperparameters} and~\ref{tab:mhc_hyperparameters} list the method-specific settings used for the main experiments.

\begin{table}[!htbp]
\centering
\small
\setlength{\tabcolsep}{3.5pt}
\renewcommand{\arraystretch}{0.92}
\caption{Shared training and data hyperparameters. Unless stated otherwise, these values are fixed across methods.}
\label{tab:global_hyperparameters}
\setlength{\tabcolsep}{5pt}
\begin{tabular}{ll}
\toprule
Hyperparameter & Value \\
\midrule
Seed & \num{34} \\
Backbone & OLMo-2-0425-1B \\
Precision & \texttt{bfloat16} \\
Dataset & \texttt{allenai/tulu-3-sft-olmo-2-mixture} \\
Input format & Chat \\
Packing & Enabled \\
Block size & \num{2048} \\
Train samples & \num{320000} (equiv.\ to batch size of 16 for \num{20000} steps) \\
Validation samples & \num{2500} \\
Test samples & \num{2500} \\
Validation/test split seed & \num{343434} \\
Optimizer & AdamW \citep{loshchilov2019decoupledweightdecayregularization} \\
Scheduler & \texttt{cosine} \\
Warmup ratio & \num{0.03} \\
Weight decay & \num{0.0} \\
Max gradient norm & \num{1.0} \\
Main training length & \num{20000} steps \\
Short diagnostic runs & \num{3000} steps \\
Evaluation interval & \num{2500} steps \\
\bottomrule
\end{tabular}
\end{table}

\begin{table}[!htbp]
\centering
\small
\setlength{\tabcolsep}{3.5pt}
\renewcommand{\arraystretch}{0.92}
\caption{Hyperparameters for standalone PEFT baselines.}
\label{tab:peft_hyperparameters}
\setlength{\tabcolsep}{4pt}
\begin{tabular}{lll}
\toprule
Method & Hyperparameter & Value \\
\midrule
LoRA & Rank $r$ & \num{8} \\
     & Alpha & \num{16} \\
     & Dropout & \num{0.05} \\
     & Bias & None \\
     & Target modules & \texttt{q,k,v,o,up,down,gate\_proj} \\
     & Learning rate & \texttt{1e-4} \\
\midrule
VeRA & Rank $r$ & \num{8} \\
     & Dropout & \num{0.05} \\
     & Projection seed & 343434 \\
     & Target modules & \texttt{q,k,v,o,up,down,gate\_proj} \\
     & Learning rate & \texttt{4e-3} \\
\midrule
$(\mathrm{IA})^{3}$ & Feedforward modules & \texttt{up,down,gate\_proj} \\
     & Bias & None \\
     & Learning rate & \texttt{3e-3} \\
\midrule
Prompt tuning & Virtual tokens & \num{20} \\
     & Initialisation text & None \\
     & Learning rate & \texttt{1e-2} \\
\midrule
Layer tuning & Train last $N$ layers & \num{1} \\
     & Train LM head & False \\
     & Train final norm & False \\
     & Learning rate & \texttt{3e-4} \\
\bottomrule
\end{tabular}
\end{table}

\begin{table}[!htbp]
\centering
\small
\setlength{\tabcolsep}{4pt}
\renewcommand{\arraystretch}{0.92}
\caption{Hyperparameters for mHC variants. Unless stated otherwise, all variants freeze the wrapped branch, use mean readout over streams, and use initialisation noise \texttt{1e-4} with residual and stream dropout set to \texttt{0}.}
\label{tab:mhc_hyperparameters}
\begin{tabular}{lll}
\toprule
Method & Hyperparameter & Value \\
\midrule
Static Sinkhorn mHC & Expansion rate $n$ & \num{4} unless varied \\
     & Residual mixer & Sinkhorn \\
     & Sinkhorn iterations & \num{20} \\
     & Sinkhorn $\epsilon$ & \texttt{1e-6} \\
     & Learning rate & \texttt{1e-3} \\
\midrule
Static mHC-lite & Expansion rate $n$ & \num{4} \\
     & Residual mixer & Birkhoff--von Neumann mixture \\
     & Learning rate & \texttt{1e-3} \\
\midrule
Static KromHC & Expansion rate $n$ & \num{4} \\
     & Residual mixer & Kronecker factors \\
     & Learning rate & \texttt{1e-3} \\
\midrule
Dynamic Sinkhorn mHC & Expansion rate $n$ & \num{4} \\
     & Residual mixer & Sinkhorn \\
     & Sinkhorn iterations & \num{20} \\
     & Sinkhorn $\epsilon$ & \texttt{1e-6} \\
     & Learning rate & \texttt{1e-4} \\
\midrule
Dynamic mHC-lite & Expansion rate $n$ & \num{4} \\
     & Residual mixer & Birkhoff-von Neumann mixture \\
     & Learning rate & \texttt{1e-4} \\
\midrule
Dynamic KromHC & Expansion rate $n$ & $\{\num{2}, \num{4}, \num{8}, \num{16}\}$ \\
     & Residual mixer & Kronecker factors \\
     & Learning rate & \texttt{1e-3} \\
\midrule
mHC$_\mathrm{identity}$ 
     & Expansion rate $n$ & $\{\num{2}, \num{4}, \num{8}, \num{16}\}$ \\
     & Residual mixer & Fixed identity \\
     & Learning rate & \texttt{1e-3} \\
\bottomrule
\end{tabular}
\end{table}

\begin{table}[!htbp]
\centering
\small
\setlength{\tabcolsep}{3.5pt}
\renewcommand{\arraystretch}{0.92}
\caption{Hyperparameters for mHC--LoRA combinations. Both runs use separate parameter groups with learning rate \texttt{1e-3} for mHC parameters and \texttt{1e-4} for LoRA parameters.}
\label{tab:combination_hyperparameters}
\setlength{\tabcolsep}{4pt}
\begin{tabular}{lll}
\toprule
Method & Hyperparameter & Value \\
\midrule
static mHC$_{n=16}$+LoRA$_{r=8}$ 
     & Backbone & OLMo-2-0425-1B \\
     & KromHC variant & Static KromHC \\
     & Expansion rate $n$ & \num{16} \\
     & $\mathbf{H}^{l}_{\mathrm{res}}$ & Learned \\
     & LoRA rank & \num{8} \\
     & LoRA alpha & \num{16} \\
     & LoRA dropout & \num{0.05} \\
     & Target modules & \texttt{q,k,v,o,up,down,gate} \\
     & mHC learning rate & \texttt{1e-3} \\
     & LoRA learning rate & \texttt{1e-4} \\
     & Training steps & \num{20000} \\
\midrule
mHC$_{\mathrm{identity}, n=2}$+LoRA$_{r=13}$
     & Backbone & OLMo-2-1124-7B \\
     & mHC variant & Dynamic KromHC \\
     & Expansion rate $n$ & \num{2} \\
     & $\mathbf{H}^{l}_{\mathrm{res}}$ & Fixed identity \\
     & LoRA rank & \num{13} \\
     & LoRA alpha & \num{16} \\
     & LoRA dropout & \num{0.05} \\
     & Target modules & \texttt{q,k,v,o,up,down,gate} \\
     & mHC learning rate & \texttt{1e-3} \\
     & LoRA learning rate & \texttt{1e-4} \\
     & Training steps & \num{20000} \\
\bottomrule
\end{tabular}
\end{table}

In case we train a combination of methods, the optimisation hyperparameters are tuned independently.
As the modular components have different search spaces and convergence qualities, the process capitalises on this \citep{oldenburg2024forecasting}: the optimiser, AdamW \citep{loshchilov2019decoupledweightdecayregularization}, is initialised with per-method learning rates and parameter groups, helping each component converge stably.

\clearpage
\section{PEFT baselines}\label{ap:PEFT_baselines}
We compare static mHC against five PEFT baselines: LoRA, VeRA, $(\mathrm{IA})^{3}$, prompt tuning, and layer tuning.
These baselines cover different adaptation mechanisms: LoRA and VeRA adapt selected weight matrices, $(\mathrm{IA})^{3}$ rescales intermediate activations, prompt tuning learns input-side soft prompts, and layer tuning unfreezes selected layers.
We also include the frozen OLMo-2 backbone without adaptation to measure the effect of finetuning relative to the base model.

PEFT adapts a pre-trained model while updating only a small number of parameters: instead of updating the full backbone, PEFT methods usually update selected components, add small modules, or scale intermediate activations \citep{han2024parameter}.
This makes them a natural point of comparison for mHC, which, instead, changes how representations are routed through the residual pathway.
Below, we introduce LoRA by \citet{hu2022lora} to contrast LoRA's parameter use with mHC. We further describe four more baselines: VeRA, $(\mathrm{IA})^{3}$, prompt tuning, and layer tuning.

LoRA is one of the most widely used PEFT methods \citep{hu2022lora, han2024parameter}.
For a frozen pre-trained weight matrix $\mathbf{W}_{0} \in \mathbb{R}^{d \times k}$, LoRA learns a low-rank update
\begin{equation}\label{eq:lora_update}
    \mathbf{W}' = \mathbf{W}_{0} + \Delta \mathbf{W} = \mathbf{W}_{0} + \mathbf{B}\mathbf{A},
\end{equation}
where $\mathbf{B} \in \mathbb{R}^{d \times r}$, $\mathbf{A} \in \mathbb{R}^{r \times k}$, $k$ and $d$ are the input and output dimensions, and $r$ is the LoRA rank, with ${r\ll\min(d,k)}$.
Finetuning updates only $\mathbf{A}$ and $\mathbf{B}$, which adds $r(d+k)$ trainable parameters per adapted matrix.

For comparison, static mHC with expansion rate $n$ adds only $n^2 + 2n$ routing parameters per layer, which depends only on the number of residual streams and not on the hidden dimension or projection size.
Since $n$ is typically tiny, this makes static mHC an extremely lightweight adaptation mechanism.
Dynamic mHC is more expressive and adds parameters that are dependent on $d$, as in Eq.~\eqref{eq:dynamic_mhc_param_count_main}.
Thus, static mHC is much smaller than LoRA, while dynamic mHC trades some of this parameter efficiency for input-dependent routing.
The difference between the methods is that LoRA adapts local weight transformations, whereas mHC adapts the routing of representations through residual streams.
The two methods can therefore be naturally combined and trained jointly.

In addition to LoRA as a PEFT baseline, we compare against VeRA, $(\mathrm{IA})^3$, prompt- and layer tuning. 
These baselines cover different adaptation mechanisms: VeRA is an ultra-light LoRA-style weight adaptation method, $(\mathrm{IA})^3$ rescales intermediate activations, prompt tuning learns input-side soft prompts, and layer tuning updates only a specific subset of layers are updated.
We explain them below in a bit more detail.

\textbf{VeRA} further reduces the trainable parameter count of LoRA-style adaptation \citep{kopiczko2024veravectorbasedrandommatrix}.
Instead of learning the low-rank matrices $\mathbf{A}$ and $\mathbf{B}$, VeRA keeps random low-rank matrices frozen and shared across layers, and only trains small scaling vectors.
For an adapted matrix with output dimension $d$ and rank $r$, this gives approximately $d + r$ trainable parameters per adapted layer, compared to $r(d+k)$ for LoRA.
VeRA is therefore a useful lightweight PEFT baseline for mHC.

\textbf{(IA)\textsuperscript{3}} learns task-specific vectors that rescale intermediate activations in attention and feed-forward layers \citep{liu2022few}.
It is from PEFT family that adapts activations rather than weight matrices. 
This is highly parameter-efficient because the learned parameters are element-wise scaling vectors rather than matrices or low-rank updates.

\textbf{Layer tuning} is a simple partial-finetuning baseline \citep{han2024parameter}.
Instead of adding new parameters, it keeps most of the pre-trained model frozen and updates only a small number of existing layers.
In our case, this means finetuning the final Transformer layer.
This baseline is useful because it tests whether updating the highest-level representations of the backbone is sufficient for adaptation.

\textbf{Prompt tuning} adapts a frozen language model by learning a small set of continuous prompt embeddings that are prepended to the input \citep{lester2021powerscaleparameterefficientprompt}.
The backbone remains frozen, and only these soft prompt vectors are updated.
For $M$ virtual tokens and hidden dimension $d$, prompt tuning adds $Md$ trainable parameters.
Unlike LoRA, DoRA, and $(\mathrm{IA})^{3}$, prompt tuning does not modify internal weights or activations directly; it changes the input context seen by the model.

Together, these methods provide baselines that act on different parts of the model.
LoRA and VeRA adapt selected weight matrices, $(\mathrm{IA})^{3}$ rescales intermediate activations, layer tuning updates a small subset of existing layers, and prompt tuning learns additional input embeddings.
mHC targets a different mechanism: it trains how representations are routed through residual streams.
This makes mHC a natural standalone PEFT candidate and a possible complement to existing PEFT methods.

Two recent finetuning methods, ReFT \citep{wu2024reftrepresentationfinetuninglanguage} and LoFiT \citep{yin2024lofitlocalizedfinetuningllm}, modify hidden state values at selected positions: ReFT learns low-rank interventions on hidden representations of a frozen model, wheras LoFiT localises task-relevant attention heads and trains offset vectors on their outputs.
Here too, mHC differs by modifying the residual routing structure itself.
Though relevant for comparison, we leave direct comparison for future work.
\section{Downstream benchmarking framework}\label{ap:benchmark_details}

To test downstream capabilities of mHC against standard PEFT baselines, all experiments are evaluated on benchmarks from the \texttt{lm-evaluation-harness} from \cite{eval-harness}. Every model configuration, as well as the untrained OLMo-2-1B model, is evaluated on the same benchmarks as the paper from \cite{xie2026mhcmanifoldconstrainedhyperconnections}.

To ensure optimal hardware utilisation on the Snellius HPC cluster, dynamic memory profiling is used via an automatic batching architecture (\texttt{batch\_size = auto}). The framework scales the hardware processing threshold to the maximum VRAM capacity of the process it is currently in, to ensure the fastest possible evaluation, without the risk of an OOM error.

As described in Table~\ref{tab:benchmark_details}, the evaluation suite adapted from \citep{xie2026mhcmanifoldconstrainedhyperconnections} spanse a diverse range of capabilities, from physical intuition and commonsense reasoning, to tasks such as multi-step logic and advance mathematical problems. This diverse layout ensures a robust assessment of each finetuning method.

\begin{table}[h]
\caption{Benchmark configurations and additional explanations/clarifications
}
\label{tab:benchmark_details}
\begin{center}
\begin{small}
\scshape 
\setlength{\tabcolsep}{3.5pt}
\resizebox{\textwidth}{!}{
\begin{tabular}{lcccc}
\toprule
Benchmark & Context Type & X-Shot Configuration & Evaluation Target & Task Metric \\
\midrule
PIQA & Physical Intuition & 0-Shot & Cloze Context Matching & Normalised Accuracy \\
HellaSwag & Commonsense Reasoning & 10-Shot & Adversarial Context Completion & Normalised Accuracy \\
MMLU & Multi-discipline Knowledge & 5-Shot & Academic Multiple-Choice & Accuracy \\
BBH & Complex Logic & 3-Shot & Algorithmic Multi-Step Parsing & Exact Match \\
GSM8K & Math Word Problems & 8-Shot & Auto-regressive CoT Generation & Exact Match \\
DROP & Discrete Reading Comp. & 3-Shot & Numerical Macro-Operation & F1-Score \\
MATH & Advanced Mathematics & 4-Shot & Symbolic Math Free-Response & Exact Match \\
TriviaQA & Factoid Knowledge & 5-Shot & Open-Ended Fact Retrieval & Exact Match \\
\bottomrule
\end{tabular}
}
\end{small}
\end{center}
\vskip -0.1in
\end{table}

\begin{table}[h]
\centering
\caption{Benchmarks used for evaluation with their categorisation.}
\vspace{0.1em}
\label{tab:downstream_benchmarks}
\footnotesize
\begin{tabular}{ll}
\toprule
\textbf{Benchmark} & \textbf{Category} \\
\midrule
BBH \citep{suzgun2022challengingbigbenchtaskschainofthought} & Complex reasoning \\
DROP \citep{dua2019dropreadingcomprehensionbenchmark} & Question answering \\
GSM8K \citep{cobbe2021trainingverifierssolvemath} & Mathematical reasoning \\
HellaSwag \citep{zellers2019hellaswagmachinereallyfinish} & Commonsense reasoning \\
MATH \citep{hendrycks2021measuringmathematicalproblemsolving} & Mathematical reasoning \\
MMLU \citep{hendrycks2021measuringmassivemultitasklanguage} & General knowledge \\
PIQA \citep{bisk2019piqareasoningphysicalcommonsense} & Commonsense reasoning \\
TriviaQA \citep{joshi2017triviaqalargescaledistantly} & Question answering \\
\bottomrule
\end{tabular}%
\end{table}
\clearpage
\section{Earlier reported findings on stability for HC and mHC}\label{ap:results_manifold}

\begin{figure}[h]
    \centering
    \includegraphics[width=1.0\textwidth]{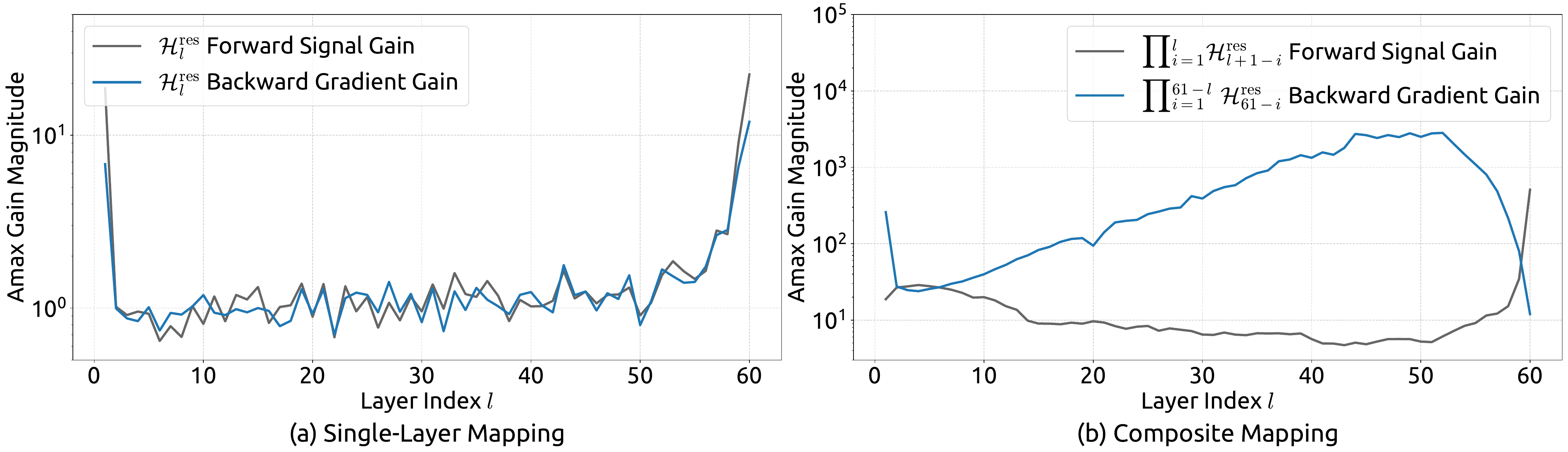}
    \caption{Propagation Instability of HC. This figure illustrates the propagation dynamics of (a) the single-layer mapping of $\mathbf{H}^{l}_{\mathrm{res}}$, here denoted as $\mathcal{H}^{\mathrm{res}}_l$, and (b) the composite mapping $\prod_{i=1}^{L-l}\mathbf{H}^{l}_{\mathrm{res}}$, here denoted as $\prod_{i=1}^{L-l}\mathcal{H}_{L-i}^{\mathrm{res}}$, within the 27B model. The layer index $l$ (x-axis) unrolls each standard Transformer block into two independent layers (Attention and FFN). The Amax Gain Magnitude (y-axis) is calculated as the maximum absolute row sum (for the forward signal) and column sum (for the backward gradient), averaged over all tokens in a selected sequence \citep{xie2026mhcmanifoldconstrainedhyperconnections}}
    \label{fig:27b_forward_backward_gain}
\end{figure}

The propagation instability of Hyper-Connections have been shown by \citet{xie2026mhcmanifoldconstrainedhyperconnections}. This can be seen in Figure \ref{fig:27b_forward_backward_gain}. 

\begin{figure}[h]
    \centering
    \includegraphics[width=1.0\textwidth]{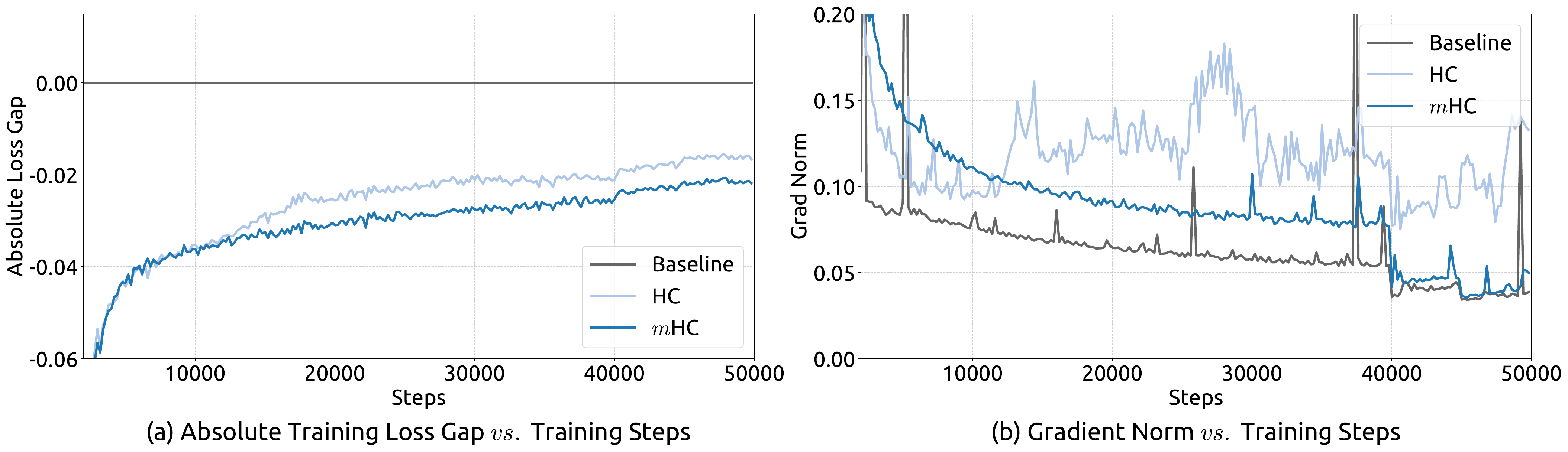}
    \caption{Training Stability of mHC. This figure illustrates (a) the absolute training loss gap of mHC and HC relative to the baseline, and (b) the gradient norm of the three methods. All experiments utilise the 27B model. The results demonstrate that mHC exhibits improved stability in terms of both loss and gradient norm \citep{xie2026mhcmanifoldconstrainedhyperconnections}
    }. 
    \label{fig:27b_loss_grad_all}
\end{figure}

On the other hand, mHC has been found to improve stability in terms of both loss and gradient norm. This can be seen in Figure \ref{fig:27b_loss_grad_all}. 

\end{document}